\documentclass[journal]{IEEEtran}
\usepackage{multirow}
\usepackage{subcaption}
\usepackage{booktabs}
\ifCLASSINFOpdf
   \usepackage[pdftex]{graphicx}
\else
\fi
\usepackage{amsmath}
\usepackage{amsfonts}
\usepackage[ruled,linesnumbered]{algorithm2e}
\usepackage{algorithmic}
\usepackage{enumitem}
\begin{document}
%
\title{DMGCRN: Dynamic Multi-Graph Convolution Recurrent Network for Traffic Forecasting}
%
%
%
\author{Yanjun~Qin,
        Yuchen~Fang,
        Haiyong~Luo,
        Fang~Zhao,
        and~Chenxing~Wang
\thanks{This work was supported in part by the National Key Research and Development Program under Grant 2018YFB0505200, the Action Plan Project of the Beijing University of Posts and Telecommunications supported by the Fundamental Research Funds for the Central Universities under Grant 2019XD-A06, the BUPT Excellent Ph.D. Students Foundation” (CX2020221), the Special Project for Youth Research and Innovation, Beijing University of Posts and Telecommunications, the Fundamental Research Funds for the Central Universities under Grant 2019PTB-011, the National Natural Science Foundation of China under Grant 61872046, the Joint Research Fund for Beijing Natural Science Foundation and Haidian Original Innovation under Grant L192004, the Key Research and Development Project from Hebei Province under Grant 19210404D and 20313701D, the Science and Technology Plan Project of Inner Mongolia Autonomous Region under Grant 2019GG328 and the Open Project of the Beijing Key Laboratory of Mobile Computing and Pervasive Device. (Corresponding author: Haiyong Luo; Fang Zhao.)}
\thanks{Yanjun Qin, Yuchen Fang, Fang Zhao, and Chenxing Wang are with the School of Computer Science (National Pilot Software Engineering School), Beijing University of Posts and Telecommunications, Beijing 100876, China (e-mail:  qinyanjun@bupt.edu.cn; fangyuchen@bupt.edu.cn; zfsse@bupt.edu.cn; wangchenxing@bupt.edu.cn).}
\thanks{Haiyong Luo is with the Research Center for Ubiquitous Computing Systems, Institute of Computing Technology Chinese Academy of Sciences, Beijing 100190, China (e-mail: yhluo@ict.ac.cn).}}

%
%

\markboth{IEEE Transactions on Neural Networks and Learning Systems}%
{Fang \MakeLowercase{\textit{et al.}}: STJLA: A Multi-Context Aware \underline{S}patio-\underline{T}emporal \underline{J}oint \underline{L}inear \underline{A}ttention Network for Traffic Forecasting}
%



\maketitle

\begin{abstract}
Traffic forecasting is a problem of intelligent transportation systems (ITS) and crucial for individuals and public agencies. Therefore, researches pay great attention to deal with the complex spatio-temporal dependencies of traffic system for accurate forecasting. However, there are two challenges: 1) Most traffic forecasting studies mainly focus on modeling correlations of neighboring sensors and ignore correlations of remote sensors, \emph{e.g.}, business districts with similar spatio-temporal patterns; 
2) Prior methods which use static adjacency matrix in graph convolutional networks (GCNs) are not enough to reflect the dynamic spatial dependence in traffic system. Moreover, fine-grained methods which use self-attention to model dynamic correlations of all sensors ignore hierarchical information in road networks and have quadratic computational complexity. In this paper, we propose a novel dynamic multi-graph convolution recurrent network (DMGCRN) to tackle above issues, which can model the spatial correlations of distance, the spatial correlations of structure, and the temporal correlations simultaneously. We not only use the distance-based graph to capture spatial information from nodes are close in distance but also construct a novel latent graph which encoded the structure correlations among roads to capture spatial information from nodes are similar in structure. Furthermore, we divide the neighbors of each sensor into coarse-grained regions, and dynamically assign different weights to each region at different times. Meanwhile, we integrate the dynamic multi-graph convolution network into the gated recurrent unit (GRU) to capture temporal dependence. Extensive experiments on three real-world traffic datasets demonstrate that our proposed algorithm outperforms state-of-the-art baselines.
\end{abstract}
\begin{IEEEkeywords}
traffic forecasting, spatial-temporal data, graph convolution network, latent graph.
\end{IEEEkeywords}

%
\IEEEpeerreviewmaketitle

\section{Introduction}
%
%
%
%
\IEEEPARstart{T}{raffic} forecasting is crucial for transportation management, public safety, and route planning in ITS \cite{dimitrakopoulos2010intelligent}. If accurate traffic forecasting in advance, it will provide some useful assistance to the transportation agencies who implement measures of controlling traffic flow, regulate route planning \cite{wu2016short}, and give early warnings to prevent the spread of COVID-19 \cite{DBLP:conf/nips/CaoWDZZHTXBTZ20}. However, traffic forecasting is very challenging, because of the complex spatio-temporal dependencies.

In the past decades, existing methods have devoted enormous efforts to capture complex spatio-temporal dependencies and forecast time-varying traffic conditions. Traditional statistical methods, such as ARIMA \cite{williams2003modeling} and VAR \cite{lu2016integrating}, are failing to capture non-linear dependencies of traffic system. The shallow machine learning methods \cite{wu2004travel,van2012short} are weak to generalize because most of them rely on hand-craft features. With the successful of deep learning in various fields intensified, early researches consider the traffic forecasting as a univariate time-series forecasting task and pay more attention to temporal dependence. The recurrent neural network (RNN) methods \cite{chung2014empirical, lv2018lc}, variants of RNN methods \cite{DBLP:conf/ssst/ChoMBB14, yao2018deep, cui2018deep}, and temporal convolution networks (TCNs) \cite{wan2019multivariate} are mainly used for traffic forecasting. Gradually, researchers found it was not enough only to capture temporal dependence. The spatial dependence of traffic system is also crucial. Therefore, some methods \cite{zhang2016dnn,zhang2017deep} utilize convolution neural networks (CNNs) to obtain spatial dependence from 2-D grid structure data (\emph{e.g.}, pictures and videos) for traffic forecasting. However, CNNs are difficult to handle the commonly used traffic data with non-Euclidean structure. Recently, graph neural networks (GNNs) \cite{DBLP:conf/iclr/KipfW17,DBLP:conf/nips/DefferrardBV16} are widely used to model non-Euclidean spatial correlations. \cite{DBLP:conf/iclr/LiYS018,DBLP:conf/ijcai/YuYZ18} combine GNNs with RNNs and TCNs to capture spatio-temporal dependencies of traffic system simultaneously. Subsequently, more GNN-based methods are proposed and they are still faced with two fundamental following challenges.

The first weakness is lacking the ability to capture the spatial dependence of remote sensors. It is not enough to only explore the spatial dependence of proximal sensors. Because of two distant sensors may be in two regions with similar functions and similar temporal patterns, \emph{i.e.}, they can bring useful information to each other \cite{lv2020temporal}. Temporal graph based methods \cite{DBLP:conf/aaai/LiZ21,fang2021spatial} use the dynamic time warping (DTW) \cite{berndt1994using} algorithm to calculate the similarity of time series of all sensors and construct a static temporal graph to get correlation information of each sensor from its temporal neighbours with a similar temporal pattern. Self-attention based methods \cite{zheng2020gman} dynamically calculate correlations of all sensors through the training phase. However, temporal graph based methods consume a lot of time, and the self-attention based methods will be negatively affected by irrelevant sensors.

The second weakness is that prior methods cannot capture the dynamic spatial dependence of traffic system. Existing GNN-based approaches \cite{DBLP:conf/iclr/LiYS018,DBLP:conf/ijcai/YuYZ18,chen2020multi,song2020spatial} are usually built on a static adjacency matrix (no matter pre-defined or self-learned) to learn spatial correlations among different sensors, even if the impact of two sensors can be changeable dynamically. Although \cite{park2020st} uses self-attention to dynamically calculate correlations of all pairs of sensors, the capability of applying to large-scale graph of it is also limited by the quadratic computation complexity.
\begin{figure}[t]
\centering
    \begin{subfigure}{0.48\linewidth}
    \includegraphics[width=\linewidth]{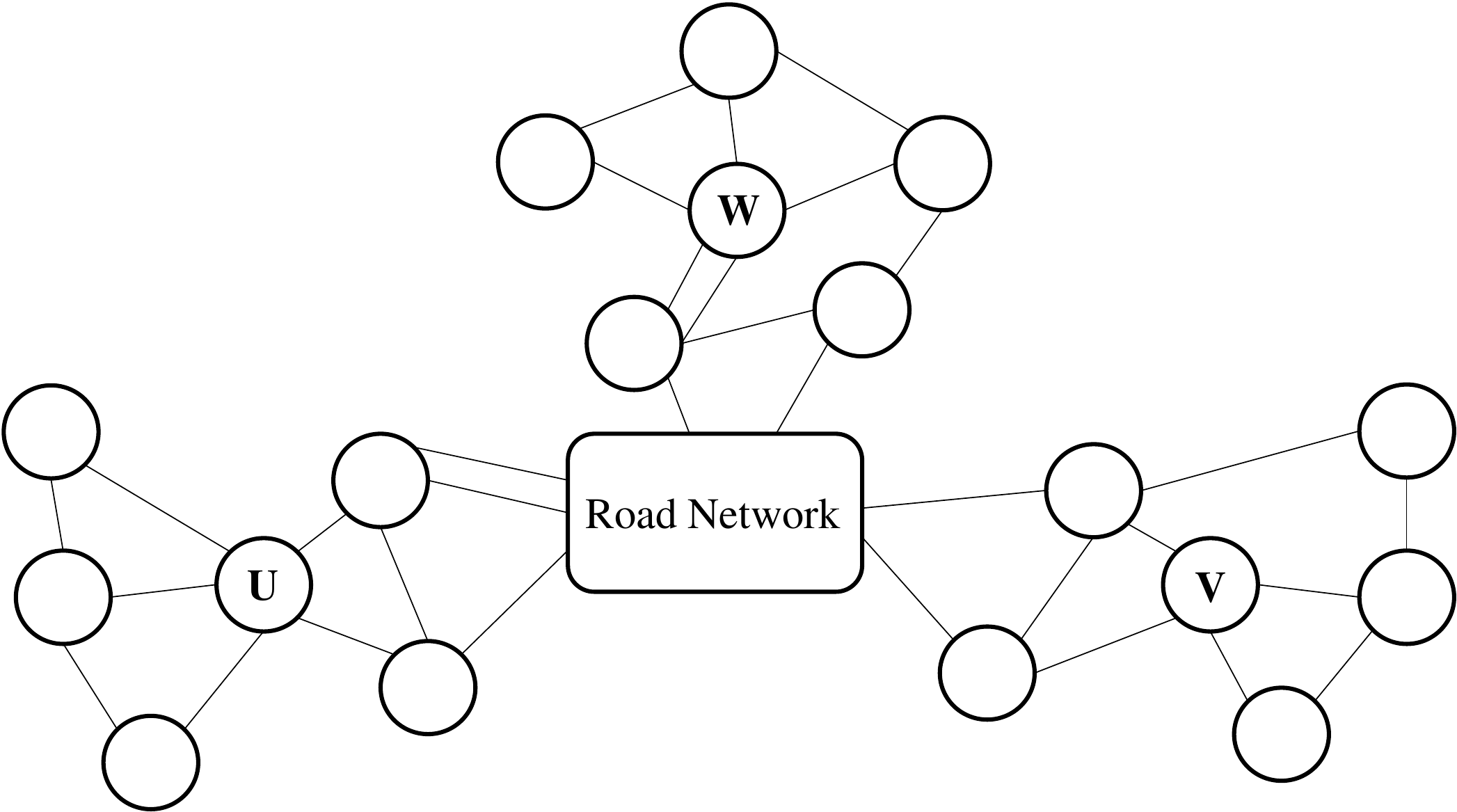}
    \caption{Road network}
    \label{rn}
  \end{subfigure}%
  \hfill
  \begin{subfigure}{0.48 \linewidth}
    \includegraphics[width=\linewidth]{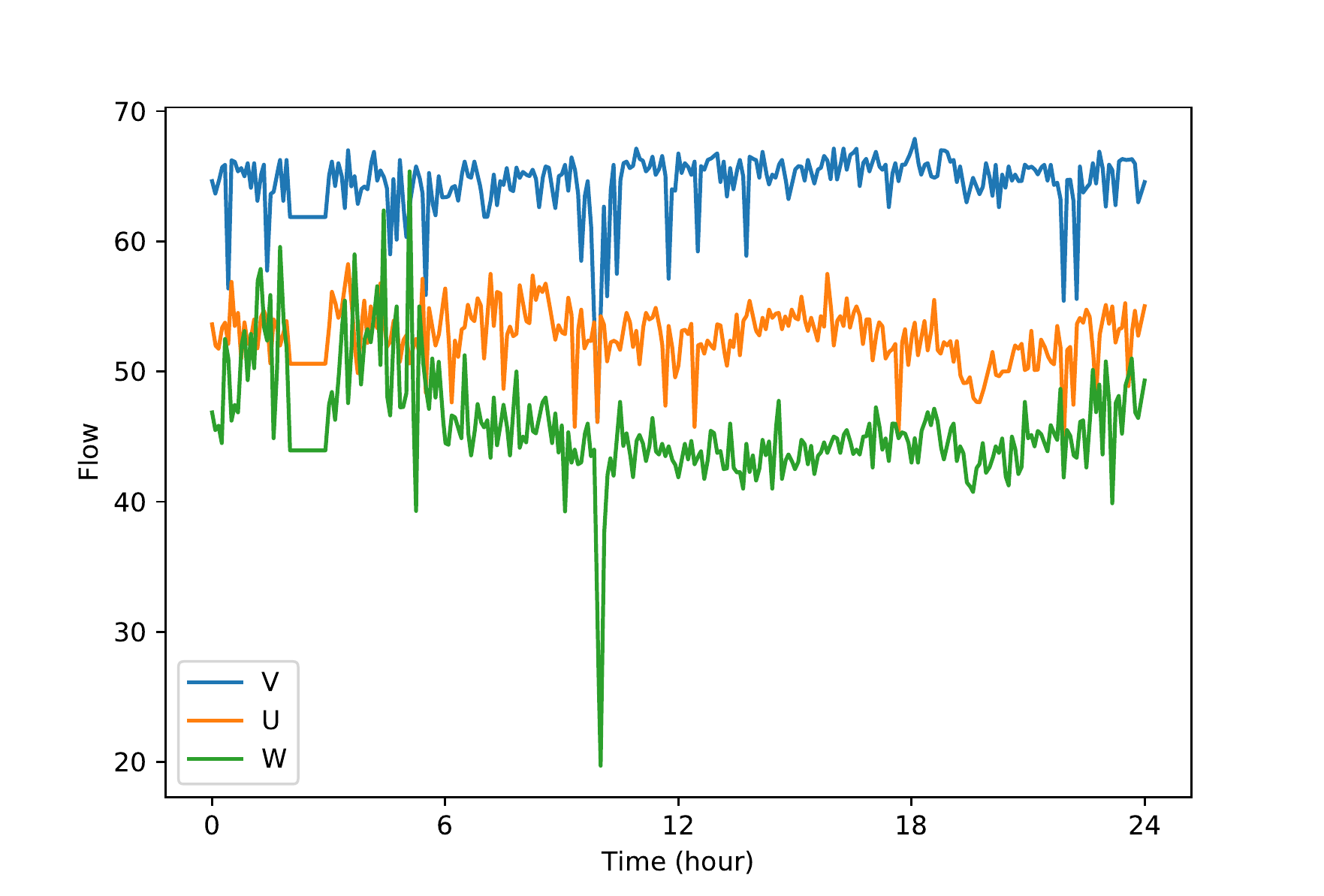}
    \caption{Time series}
    \label{ts}
  \end{subfigure}%
  \caption{(a): visualizes a road network, the spatial structure of the sensors u, v, and w in the road network is similar. (b): shows the time series of sensors u, v, and w.}
  \label{exa}
\end{figure}

In this paper, we propose a novel \underline{d}ynamic \underline{m}ulti-\underline{g}raph \underline{c}onvolution \underline{r}ecurrent \underline{n}etwork (DMGCRN) to address above issues, which integrates the dynamic multi-graph convolution network into GRU to capture spatio-temporal dependencies simultaneously. In DMGCRN, we design a novel latent representation based latent graph and the shortest path distance based distance graph to collaboratively discover spatial correlations of sensors are close in distance and similar in structure. Because we observe regions with the same functions have similar structures and similar temporal patterns. As shown in Figure \ref{exa}, three sensors that are far apart in the road network are in similar structures, and the trend of their speed within a day is like each other. We use the latent graph to replace the temporal graph in prior models to obtain remote information because it is more resource-efficient. Besides, the change of speed of sensors in the road network can not only be modeled as a fine-grained sensor relationship but also as a message passing from different regions to the sensors \cite{guo2021hierarchical}. Therefore, we split neighbors of each sensor into different regions according to the relationship between the sensor and neighbors in the coordinate, then we utilize the graph convolution network and the attention mechanism to capture the diverse spatial dependencies in different regions and dynamically assign weights to regions, respectively. The major contributions of our work can be summarized as follows:
\renewcommand\labelitemi{$\bullet$}
\begin{itemize}
    \item A dynamic graph convolution network is proposed. Different from existing fine-grained methods, our method dynamically captures coarse-grained region correlations.
    \item A latent graph is constructed by the similarity of latent representation of sensors in road networks, which can discover similar temporal patterns of remote sensors.
    \item We conduct extensive experiments on three real-world traffic datasets, METR-LA, PEMS03, and PEMS05. Our model consistently outperforms all the baseline methods.
\end{itemize}

\section{Related Works}
\subsection{Traffic Forecasting}
Traffic forecasting has been studied for decades. Researchers use traditional time series methods (\emph{e.g.}, auto-regressive integrated moving average (ARIMA) \cite{williams2003modeling}) and machine learning models (\emph{e.g.}, support vector regression (SVR) \cite{wu2004travel}, Kalman filtering \cite{lippi2013short}) in the early year. However, traditional time series models usually rely on the stationary assumption and machine learning models need to extract hand-craft features, which require a lot of prior knowledge. As the development of deep learning, \cite{hochreiter1997long,chung2014empirical} utilize deep learning models for traffic forecasting, they only consider the temporal dependence. \cite{zhang2017deep,zhang2016dnn} use convolution neural networks (CNNs) to capture spatial correlations of images. STGCN and DCRNN \cite{DBLP:conf/ijcai/YuYZ18,DBLP:conf/iclr/LiYS018} further leverage the GCN to capture spatial dependence in non-Euclidean structure but limited by the local receptive field of pre-defined graph. Subsequent methods, such as STFGNN \cite{DBLP:conf/aaai/LiZ21}, use DTW algorithm to get global spatial information by calculating the historical traffic pattern correlation. They are limited by the slow calculation speed of DTW in long sequences. Although \cite{DBLP:conf/ijcai/WuPLJZ19,DBLP:conf/nips/0001YL0020,wu2020connecting} learn a more complete graph through back-propagation and \cite{zheng2020gman} use the self-attention mechanism instead of GCN to extract the global spatial dependence. They are negatively affected by irrelevant sensors.
\subsection{Graph Convolution Network}
Recently, Graph convolution networks (GCNs) have been a widely applied graph structure data analysis method. Graph convolution networks are categorized as spatial convolution and spectral convolution approaches. Spectral convolution approaches define graph convolution operation in the spectral domain. For example, \cite{DBLP:journals/corr/BrunaZSL13} proposes a spectral graph convolution operation, which is defined in the Fourier domain by computing eigen-decomposition of the graph Laplacian. ChebNet \cite{hammond2011wavelets} suggests that the convolution operation can be further approximated by the K-th order polynomial of graph Laplacian. Kipf \emph{et al.} \cite{DBLP:conf/nips/DefferrardBV16,DBLP:conf/iclr/KipfW17} propose a simpler GCN which limits the layer-wise convolution operation to $K = 1$ and introduces a re-normalization trick. Spatial convolution methods usually define graph convolution operations directly on the graph. For instance, DCNN \cite{atwood2016diffusion} proposes the diffusion convolution neural network represents the diffusion process on the graph. GraphSAGE \cite{DBLP:conf/nips/HamiltonYL17} generates embeddings by sampling and aggregating features from the neighborhood of nodes. Graph attention networks \cite{DBLP:conf/iclr/VelickovicCCRLB18} incorporates the attention mechanism to assign different weights to a neighborhood.
\begin{figure*}[t]
    \centering
    \includegraphics[width=1.0\linewidth]{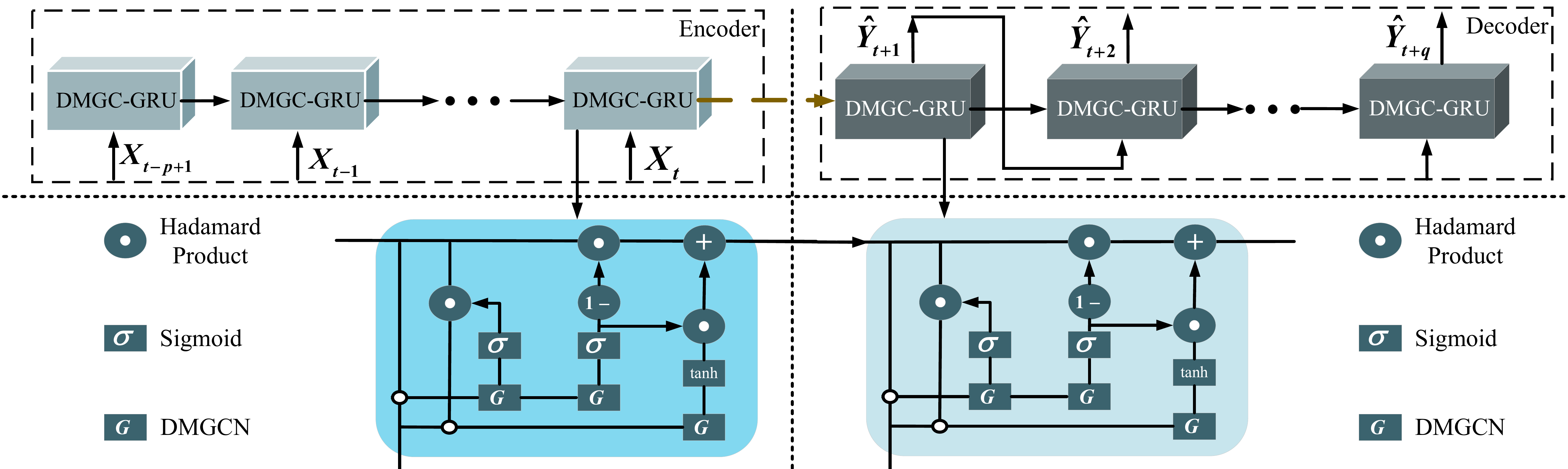}
    \caption{The upper part is the overall architecture of the DMGCRN, and the lower part is the enlarged DMGC-GRU unit}
    \label{dmgcrn}
\end{figure*}
\subsection{Attention mechanism}
The addictive attention \cite{DBLP:journals/corr/BahdanauCB14} proposed to improve the word alignment in the translation task. Then, \cite{DBLP:conf/emnlp/LuongPM15} proposes the widely used general and dot-product attention. The popular dot-product based self-attention has been proposed in Transformer \cite{DBLP:conf/nips/VaswaniSPUJGKP17} and has achieved brilliant success in the NLP \cite{liddy2001natural} and CV \cite{forsyth2011computer}. In this paper, we use the attention mechanism to distinguish the importance of regions for each sensor.
\section{Preliminaries}
\subsection{Problem Definition}
The goal of the traffic forecasting task is to predict future traffic data  (\emph{e.g.}, traffic volume, traffic speed, \emph{etc.}) based on the historical data recorded by sensors in the road network. The topology of the real road network is expressed as a weighted directed graph $\mathcal{G}=(\mathcal{V},\mathcal{E},\mathcal{A})$, where $\vert\mathcal{V}\vert=N$ is a set of nodes and each node represents a sensor in the road network. $\mathcal{E}$ is a set of edges and $\mathcal{A}\in\mathbb{R}^{N\times N}$ is the adjacency matrix of $\mathcal{E}$. The traffic data in the road network at time slice $t$ is represented as a graph signal $X_t\in \mathbb{R}^{N\times F}$, where $F$ is the feature dimension of the traffic data. Given historical traffic data of the whole traffic network over past $T^h$ time slices, denoted as $\mathcal{X} = [X_1,X_2,...,X_{T^h}]\in \mathbb{R}^{T^h\times N \times F}$, the aim of traffic forecasting problem is learn a function $f(\cdot)$, which is used to predict the next $T^p$ time slices traffic data, denoted as:
\begin{equation}
    [X_1,X_2,...,X_{T^h};\mathcal{G}]\mathop{\to}_{f}[\hat{Y}_1,\hat{Y}_2,...,\hat{Y}_{T^p}]\in\mathbb{R}^{T^p\times N \times F}\quad .
\end{equation}
\section{Methodology}
Figure \ref{dmgcrn} elaborates on the general framework of our model. DMGCRN adopts an encoder-decoder architecture to capture historical spatio-temporal correlations and forecast future traffic series step-by-step. In each step, we use the DMGC-GRU to capture spatio-temporal dependencies simultaneously, which replaces the linear operator in the GRU with our DMGCN. The details of DMGCRN are illustrated in the following.

\subsection{Dynamic Multi-Graph Convolution Network}
\begin{figure}
    \centering
    \includegraphics[width=1.0\linewidth]{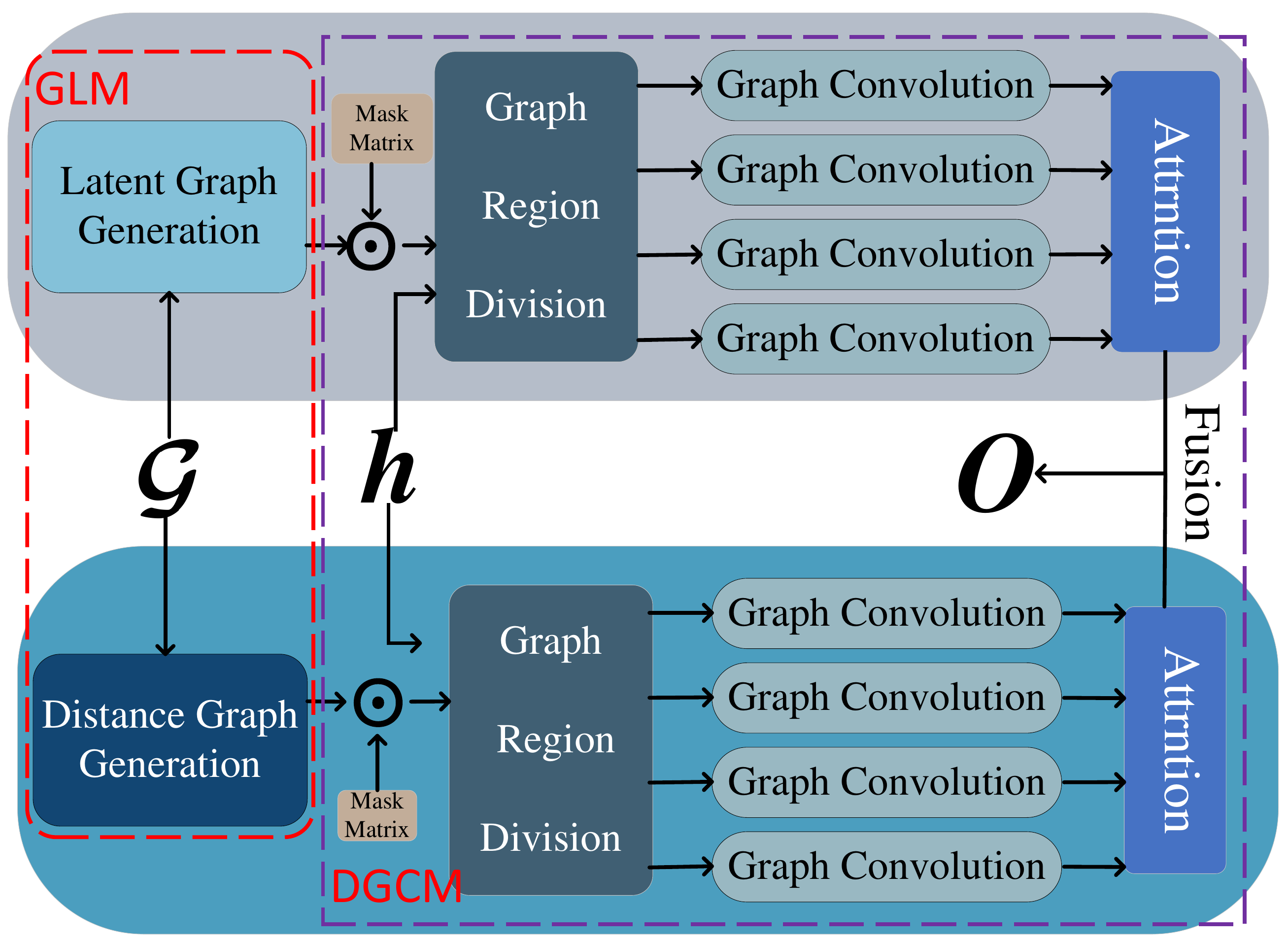}
    \caption{The architecture of the DMGCN. GLM and DGCM denotes the graph learning module and the dynmaic graph convolution module, respectively. Besides, $h$, $O$, $\mathcal{G}$, and $\odot$ indicates the input of the DMGCN, the output of the DMGCN, road networks, and the dot-product operation.}
    \label{dmgcn}
\end{figure}

To discover hidden spatial correlations among sensors, as shown in Figure \ref{dmgcn}, DMGCN uses a graph learning module to construct the distance graph adjacency matrix based on the shortest path distance and the latent graph adjacency matrix based on the latent representation of sensors, which are later used as inputs to two individual dynamic graph convolution modules. Moreover, the dynamic graph convolution module uses the graph region division layer, graph convolution layer, and attention layer in turn to dynamically capture spatial dependence. At the end of DMGCN, we utilize a fusion module to fuse the results of the two graphs.
\subsubsection{Graph Learning Module}
\paragraph{Distance Graph Generation Layer} We get a directed weighted graph according to the actual road network follow the previous study \cite{song2020spatial}, and we use the Dijkstra algorithm \cite{dijkstra1959note} to calculate the shortest path distance $D^{spd}$ between all pairs of nodes to obtain an augmented fully graph. Finally, we use threshold Gaussian kernel to build the shortest path distance based graph adjacency matrix $\mathcal{A}^d\in\mathbb{R}^{N\times N}$:
\begin{equation}
    \mathcal{A}^d_{i,j}=\begin{cases} 1& \text{, $i\neq j$   and $exp(-\frac{D^{spd^2}_{i,j}}{\delta})\ge\epsilon$}\\0& \text{, otherwise} \end{cases}\quad,
\end{equation}
$\delta$ denotes the standard deviation and $\epsilon$ is the thresholds to control the distribution sparsity of $\mathcal{A}^d$.
\paragraph{Latent Graph Generation Layer} Existing frameworks typically only utilize the graph of the actual road network to model spatial correlations, which ignores the spatial dependence of remote sensors for traffic forecasting. The way of increasing the number of layers of GCN can expand receptive field and receive traffic information from distant nodes, which also leads to the over-smoothing phenomenon \cite{chen2020simple} and increases the resource consumption during training. Inspired by methods of constructing a temporal graph by calculating the similarity of time series of sensors using the DTW to capture remote spatial correlations, we construct a novel latent graph to replace temporal graph to capture remote spatial correlations by calculating the similarity of the local structure of sensors using the struc2vec \cite{ribeiro2017struc2vec}. The struc2vec is a graph representation learning algorithm. Specifically, we first use struc2vec on the road network based graph adjacency matrix to learn the latent representation $E\in\mathbb{R}^{N\times d^e}$ of all nodes with local spatial structure information. Because struc2vec can create particular hyperbolic spaces that preserve hierarchies and local structures in a graph \cite{DBLP:conf/iclr/PeiWCLY20}, we use the distance metric in the hyperbolic space to calculate the similarity of the latent representation between all sensors. For instance, the calculation of the similarity $D^{lr}_{i,j}$ between sensor $i$ and $j$ is:
\begin{equation}
    D^{lr}_{i,j}=arcosh(1+2\frac{\Vert E_i-E_j\Vert^2}{(1-\Vert E_i\Vert^2)(1-\Vert E_j\Vert^2)})\quad ,
\end{equation}
where $E_i,E_j\in\mathbb{R}^{d^e}$ are the latent representations of sensor $i$ and $j$, and $d^e$ denotes the embedding dimension of struc2vec. Besides, the $arcosh(x)=ln(x+\sqrt{x^2-1})$ is an inverse hyperbolic cosine function.

Similar to the build of shortest path distance based adjacency matrix $\mathcal{A}^d$, we use threshold Gaussian kernel to build the latent representation based adjacency matrix $\mathcal{A}^l\in\mathbb{R}^{N\times N}$:
\begin{equation}
    \mathcal{A}^l_{i,j}=\begin{cases} 1& \text{, $i\neq j$   and $exp(-\frac{D^{lr^2}_{i,j}}{\delta})\ge\epsilon$}\\0& \text{, otherwise} \end{cases}\quad.
\end{equation}
\subsubsection{Dynamic Graph Convolution Module} \label{DGCN}
In this section, we use two completely consistent dynamic graph convolution modules to capture the spatial dependence in the road network, except that graphs are different. For brevity, we only introduce the process of the latent graph.
\paragraph{Mask Mechanism} In order to distinguish the importance of different neighbor nodes in the latent graph, we design a mask mechanism to adjust the edge weights of the graph. Specifically, we use a learnable parameter matrix $M\in\mathbb{R}^{N\times N}$ to multiply the latent graph adjacency matrix $\mathcal{A}^l\in\mathbb{R}^{N\times N}$, and then change the value of the parameter matrix through back-propagation to change the value of the adjacency matrix:
\begin{equation}
    \mathcal{\bar{A}}^l=M\odot \mathcal{A}^l\quad ,
\end{equation}
where $\mathcal{\bar{A}}^l$ denotes the learnable latent graph adjacency matrix.
\paragraph{Graph Region Division Layer} The previous methods use a static adjacency matrix to model the spatial correlations and can not reflect the dynamically evolving road network. However, using attention mechanism to fine-grained model the spatial correlations between nodes in each time slice requires a lot of computing resources. The transportation system contains hierarchical spatial correlations, \emph{i.e.}, in addition to fine-grained node-level correlations, there are also coarse-grained region-level correlations. The dynamic modeling of the region can greatly reduce the complexity of the model and increase the running speed. Different from the existing method, which absolutely divides nodes into different regions \cite{guo2021hierarchical}, we divide neighbors of each node into different regions according to the relative relationship between nodes. For the distance graph, we use the relative position of the sensor's location (\emph{i.e.}, latitude and longitude) in the 2-D Euclidean space to divide the neighbors of each sensor into four different regions. Therefore, we use an embedding space of dimension 2 in struc2vec to divide the neighbors of each sensor in latent graph into four different regions according to their relative position of the sensor's latent representation in the hyperbolic space. Concretely, we select edges from different regions of each sensor and save all sensors edge of each region into an adjacency matrix individually. The adjacency matrix of the latent graph which saves all regions' edge is $\tilde{\mathcal{A}}^l\in\mathbb{R}^{4\times N\times N}$.
\paragraph{Graph Convolution Layer} To capture spatial dependence of the non-Euclidean structure efficiently and accurately, we use the graph convolution network as proposed in \cite{DBLP:conf/iclr/KipfW17}, where node representations are computed by aggregating messages from direct neighbors. For the hidden input $h\in\mathbb{R}^{N\times d^h}$ of each time slice, the formulate of information propagates on the latent region graph $\tilde{\mathcal{A}}^l$ is:
\begin{equation}
\tilde{h}^l=\sigma(\mathop{\Vert}^4_{i=1}(D^{l^{-\frac{1}{2}}}_i(\tilde{\mathcal{A}}^l_i+I)D^{l^{-\frac{1}{2}}}_i)hW)\quad ,
\end{equation}
where $d^h$ is the feature dimension of input, $d^h=d+F$ and $d^h=2d$ in DMGCNs where in the first layer of GRU and other layers. Besides, $d$ is the feature dimension of GRU units. $\tilde{h}^l\in\mathbb{R}^{4\times N\times d}$ is the output updated by edges of four regions in the latent graph. $\Vert$ denotes the concatenate operation. $W\in\mathbb{R}^{d^h\times d}$ is a shared parameter in graph convolution, which reduces the number of parameters and enhances the generalization ability of the model. In Equation 6, the identity matrix $I\in\mathbb{R}^{N\times N}$ is added to the adjacency matrix $\tilde{\mathcal{A}}^l_i$ to obtain self-loops for each node, and the resultant matrix is normalized using the diagonal degree matrix $D_{i,j,j}^l=\sum\nolimits_k\tilde{\mathcal{A}}^l_{i,j,k}+I_{j,j}$.
\paragraph{Attention Layer}
After the graph convolution, we get the output of different regions. Then we utilize the attention mechanism to aggregate spatial information from different regions dynamically. For sensor $i$, we first adopt linear transformation to reduce the feature dimension of all regions hidden representation $\tilde{h}_{i}^l\in\mathbb{R}^{4\times d}$ to 1, and then apply the $softmax$ activation function on it to derive the attention coefficient $\alpha_i\in\mathbb{R}^{4\times 1}$, which denotes the correlations between the final output and regions. Finally, the output $O^l_i\in\mathbb{R}^{d}$ can be obtained by the dot-product between with attention coefficients and the corresponding region outputs. The operation can be represented as:
\begin{equation}
    \begin{split}
        e^{l}_i &= W^{l}_{2}(ReLU(W^{l}_{1}h^{l}_i+b^{l}_{1}))+b^{l}_{2}\quad ,\\
        \alpha^{l}_i &= softmax(e^l_i),\quad O^{l}_i= \alpha_i^l\tilde{h}^l_i\quad ,\\
    \end{split}
\end{equation}
where $ReLU$ is the activation function. $W_{1}^{l}\in\mathbb{R}^{d\times d}$, $W_{2}^{l}\in\mathbb{R}^{d\times 1}$ are learnable parameters.  $b_{1}^{l}\in\mathbb{R}^{d}$, $b_{2}^{l}\in\mathbb{R}$ are bias terms. $O^{l}\in\mathbb{R}^{N\times d}$ is the output of dynamic graph convolution with the latent graph. Similarity, we derive the output $O^{d}\in\mathbb{R}^{N\times d}$ of dynamic graph convolution with the distance graph.
\subsubsection{Fusion Module}
In this part, we make the dimension of hidden input consistent with the outputs of the attention layer by linear transformation and then fuse the outputs of three components. The fusion of DMGCN is formulated as:
\begin{equation}
    O = W_2^h(W_1^hh)+W^lO^{l}+W^dO^{d}\quad ,
\end{equation}
$W_1^h\in\mathbb{R}^{d^h\times d}$ and $W_2^h,W^l,W^d\in\mathbb{R}^{d\times d}$ are learnable parameters, reﬂecting the impact of each components on the DMGCN. $O\in\mathbb{R}^{N\times d}$ is the output of DMGCN.
\subsection{Temporal Dependence Modeling}
RNN, LSTM and GRU are the common methods to model the temporal dependence. We choose GRU to learn temporal correlations because it is a simple and powerful variant of RNN. We replace the linear operator in GRU with the DMGCN and get DMGC-GRU. Formulated as:
\begin{equation}
\begin{split}
    U_{t}&=\sigma(\mathcal{G}([X_{t}\Vert H_{t-1}];\Theta_U))\quad ,\\R_{t}&=\sigma(\mathcal{G}([X_{t}\Vert H_{t-1}];\Theta_R))\quad ,\\C_{t}&=tanh(\mathcal{G}([X_{t}\Vert (R_{t}\odot H_{t-1})];\Theta_C))\quad ,\\H_{t}&=U_{t}\odot H_{t-1}+(1-U_{t})\odot C_{t}\quad ,
\end{split}
\end{equation}
where $\mathcal{G}(\cdot)$ denotes the DMGCN. $X_{t}$ is the input of DMGC-GRU at timestep $t$, $X_{t}\in\mathbb{R}^{N\times F}$ and $X_{t}\in\mathbb{R}^{N\times d}$ in the first layer and other layers. $H_{t-1}\in\mathbb{R}^{N\times d}$ and $H_{t}\in\mathbb{R}^{N\times d}$ are the previous hidden state and output hidden state of DMGC-GRU at timestep $t$, respectively. $U_{t}$ denotes the update gate, $R_{t}$ denotes the reset gate, and $C_{t}$ denotes the candidate activation vector at timestep $t$. $\sigma$ is the sigmoid activation function and $\odot$ is the hadamard product. $\Theta_U$, $\Theta_R$, and $\Theta_C$ are parameters for the corresponding filters, respectively. Besides, the scheduled sampling \cite{DBLP:conf/nips/BengioVJS15} is applied in our DMGC-GRU for better generalization.

  

\subsection{Loss Function}
DMGCRN can be trained end-to-end via back-propagation by
minimizing the L1 Loss between the predicted values and ground truths:
\begin{equation}
    \mathcal{L}(\Theta) = \frac{1}{N\times T^p}\sum_{t=1}^{T^p}\sum_{i=1}^N\vert X_{t,i}-\hat{Y}_{t,i}\vert\quad ,
\end{equation}
where $\hat{Y}$ is the prediction value, $X$ denotes the ground truth, and $\Theta$ represents all trainable parameters. The detailed training procedure of DMGCRN is summarized in Algorithm 1.
\begin{algorithm}[t]
\caption{Training algorithm of DMGCRN.}
\KwData{Road network graph $\mathcal{G}=(\mathcal{V},\mathcal{E},\mathcal{A})$, traffic tensor $X\in\mathbb{R}^{T\times N\times F}$ of train set, initialized DMGC-GRU unit $f_{dg}(\cdot)$, pre-defined function for scheduled sampling $f_{ss}(\cdot)$, learning rate $\gamma$, learnable parameters $\Theta$.}
Employ dijkstra and struc2vec algorithm to get the distance graph $\mathcal{A}^d$ and the latent graph $\mathcal{A}^l$.\\
Initialize all learnable parameters $\Theta$ in DMGCRN.\\
set $iter=1$.\\
\Repeat{met model stop criteria}{
initialize hidden state $H_0$ and $iter=iter+1$.\\
randomly select a batch of input sample $\mathcal{X}\in\mathbb{R}^{B\times T^h\times N\times F}$ from $X$.\\
randomly select a batch of label sample $\mathcal{Y}\in\mathbb{R}^{B\times T^p\times N\times F}$ from $X$.\\
\For{$t\leftarrow 1$ \KwTo $T^h$}{
    compute $H_{t}=f_{dg}(\mathcal{X}_{t},H_{t-1},\mathcal{A}^d,\mathcal{A}^l,\Theta)$.\\
}
initialize traffic signal $\mathcal{\hat{Y}}_0\in\mathbb{R}^{B\times N\times F}$ as a zero vector for decoder.\\
\For{$t\leftarrow 1$ \KwTo $T^p$}{
    randomly select a number $c\sim \mathcal{U}(0,1)$.\\
    \eIf{$c<f_{ss}(iter)$}{
        $\mathcal{Y}_{t}=\mathcal{X}_{t}$
    }
    {
        $\mathcal{Y}_{t}=\mathcal{\hat{Y}}_{t-1}$
    }
    compute $\mathcal{\hat{Y}}_t,H_{T^h+t}=f_{dg}(\mathcal{Y}_{t},H_{T^h+t-1},\mathcal{A}^d,\mathcal{A}^l,\Theta)$.\\
}
compute the loss of the Equation 10.\\
compute stochastic gradient of $\Theta$ according to loss.\\
update $\Theta$ according to gradient and learning rate $\gamma$.\\
}
\KwResult{Learned DMGCRN model.}
\end{algorithm}
\begin{table}[t]
\centering
\caption{Statistics of datasets.}
\label{dataset}
\resizebox{1.0\linewidth}{!}{\begin{tabular}{lcccc}  
\toprule
Datasets  & \#Nodes & \#Edges & \#TimeSteps & TimeSpan \\
\midrule
METR-LA & 207 & 1515 & 34272 & 2012/5-2012/6   \\
PEMS03   & 358 & 547 & 26208 & 2018/9-2018/11      \\
PEMS05   & 205 & 269 & 18144 & 2020/6-2020/8\\
\bottomrule
\end{tabular}}
\end{table}

\section{Experiments}
\subsection{Datasets Description}
We evaluate the performance of DMGCRN on three real-world traffic network datasets. METR-LA released by \cite{DBLP:conf/iclr/LiYS018}, PEMS03 released by \cite{song2020spatial}, and PEMS05 constructed by us. The statistical information is shown in Table \ref{dataset}.
\paragraph{METR-LA}
This traffic dataset contains traffic speed information collected from loop detectors on the highway of Los Angeles County, with 207 sensors and collects 4 months of data ranging from Mar 1st 2012 to Jun 30th 2012.
\paragraph{PEMS03}
This traffic dataset is collected by California Transportation Agencies (CalTrans) Performance Measurement System (PeMS) in 3 districts, with 358 sensors and for 2 months of data ranging from Jan 1st 2018 to May 31th 2018.
\paragraph{PEMS05}
This traffic dataset is collected by PeMS in 5 districts, with 205 sensors and collects 2 months of data ranging from Jan 1st 2020 to May 31th 2020.

We aggregate these traffic datasets into 5-minute windows. Besides, inputs are normalized by z-score normalization.
\subsection{Metrics}
We adopt three commonly used metrics, including Mean Absolute Errors (MAE), Mean Absolute Percentage Errors (MAPE), and Root Mean Squared Errors (RMSE) to measure the performance of different methods. As follows:
\begin{equation}
    \begin{split}
        MAE&=\frac{1}{N\times T^p}\sum_{t=1}^{T^p}\sum_{i=1}^N\vert X_{t,i}-\hat{Y}_{t,i}\vert\quad ,\\
        RMSE&=\sqrt{\frac{1}{N\times T^p}\sum_{t=1}^{T^p}\sum_{i=1}^N( X_{t,i}-\hat{Y}_{t,i})^2}\quad ,\\
        MAPE&=\frac{1}{N\times T^p}\sum_{t=1}^{T^p}\sum_{i=1}^N \frac{\vert X_{t,i}-\hat{Y}_{t,i}\vert}{X_{t,i}}\quad .
    \end{split}
\end{equation}
\subsection{Baselines}
The description of baselines is as follows:
\begin{itemize}
\item HA \cite{williams2003modeling}: Historical average, which uses the average of the historical series to predict the future traffic value.
\item SVR \cite{wu2004travel}: Linear support vector regression uses a support vector machine to do regression on the traffic sequence. This framework ignores the spatial dependence on traffic forecasting. We use the rbf kernel in SVR for iterative multi-step prediction.
\item FC-LSTM \cite{sutskever2014sequence}: The Encoder-decoder framework implemented by using three LSTM layers with 64 hidden units.
\item DCRNN \cite{DBLP:conf/iclr/LiYS018}: Diffusion convolution recurrent neural network, which combines graph convolution networks with recurrent neural networks in an encoder-decoder architecture. The most limit of this model is that it only relies on the road network. We use the parameters in the original paper for experiments.
\item STGCN \cite{DBLP:conf/ijcai/YuYZ18}: Spatial-temporal graph convolution network, which combines graph convolution with 1D CNN. The spatio-temporal convolution comprises graph convolution layers and 1-D causal convolutions to model spatial and temporal dependencies. The 1D convolution is lack of capacity for modeling long-term dependence. We keep settings in experiments consistent with original paper.
\item Graph WaveNet \cite{DBLP:conf/ijcai/WuPLJZ19}: It uses a self-adaptive graph in graph convolution networks to capture spatial dependence and uses 1D dilated convolution to capture temporal dependence. All settings follow the original paper.
\item AGCRN \cite{DBLP:conf/nips/0001YL0020}: It uses a self-adaptive graph in graph convolution networks to capture the hidden spatial dependence and uses GRU to capture the temporal dependence. We fine-tuned node embedding based on each dataset.
\item STFGNN \cite{DBLP:conf/aaai/LiZ21}: It proposes a spatio-temporal fusion graph to capture spatio-temporal dependencies synchronously. Besides, it designs a learnable matrix to adjust weights of adjacency matrix through back-propagation.
\item GMAN \cite{zheng2020gman}: Graph multi-attention network is a Transformer-like model that applies the self-attention mechanism to spatio-temporal dimensions to capture global spatio-temporal dependencies dynamically. However, it is limited by the quadratic problem of the self-attention, which consumes a lot of time and memory.
\end{itemize}
\begin{table}[t]
\centering
\caption{Experiment hyperparameter settings}
\label{param}
\resizebox{0.8\linewidth}{!}{\begin{tabular}{lc}
\toprule
Hyperparameter description & Value \\
\midrule
Batchsize & 32\\
Number of training epochs & 80\\
Learning rate & 0.001\\
Learning decay rate & 0.2\\
Learning decay epoch & [30,40,50]\\
Optimizer & Adam\\
GRU layers & 2\\
Dimensions & 64\\
Embedding dimensions & 2\\
\bottomrule
\end{tabular}}
\end{table}
\subsection{Experiment Settings}
We split the METR-LA dataset with a ratio $7:1:2$ into training sets, validation sets and test sets, as well as we split the PEMS03 and PEMS05 datasets with a ratio $6:2:2$ into training sets, validation sets and test sets in chronological order. All experiments are run on a CentOS Linux server (CPU: Intel(R) Xeon(R) Gold 6132 CPU @ 2.60GHz, GPU: Tesla-V100). We use the historical one hour data (\emph{i.e.}, 12 timesteps) to forecast traffic conditions in the next hour.
\paragraph{Our Model}The hyperparameters of our model are shown in Table \ref{param}. We implement our model using PyTorch.
\begin{table*}[t]
\aboverulesep=0ex
\belowrulesep=0ex
\centering
\caption{Traffic speed prediction performance comparison of DMGCRN and baselines on METR-LA. }
\label{metr}
\resizebox{\textwidth}{!}{
\begin{tabular}{c|l|ccc|ccc|ccc} 
\toprule
\multirow{2}{*}{Datasets} & \multirow{2}{*}{Methods} & \multicolumn{3}{c|}{15 min} & \multicolumn{3}{c|}{30 min} & \multicolumn{3}{c}{1 hour} \\
\cmidrule{3-11}
    &  & MAE & RMSE & MAPE (\%) & MAE & RMSE & MAPE (\%) & MAE & RMSE & MAPE (\%) \\
\midrule
\multirow{10}{*}{METR-LA}
& HA & 4.80 & 10.02  & 11.74 & 5.48  & 11.46 & 13.52 & 7.00 & 13.91 & 17.56\\
& SVR & 3.61 & 8.12  & 8.96 & 4.59 & 10.33 & 11.79 & 6.05 & 13.04 & 16.16\\
& FC-LSTM & 3.52 & 7.97 & 9.04 & 4.53  & 10.29 & 12.02 & 6.05 & 12.98 & 16.63\\
\cmidrule{2-11}
& DCRNN & 2.80 & 5.40 & 7.23 & 3.18  & 6.41 & 8.65 & 3.63 & 7.49 & 10.40\\
& STGCN & 2.81 & 5.57 & 7.70 & 3.41 & 7.43 & 9.27 & 4.79 & 9.71 & 12.08\\
\cmidrule{2-11}
& Graph WaveNet & 2.70 & 5.18 & 7.02 & 3.11 & 6.23 & 8.44 & 3.57 & 7.39 & 10.05\\
& AGCRN & 3.78 & 8.48 & 9.67 & 4.80 & 12.14 & 10.33 & 6.16 & 14.94 & 12.69\\
& STFGNN & 3.26 & 7.43  & 8.04 & 4.03  & 9.44 & 10.22 & 5.02 & 11.62 & 13.03\\
\cmidrule{2-11}
& GMAN & 2.78 & 5.49  & 7.37 & 3.09  & 6.29 & 8.44 & 3.46 & 7.38 & 9.94\\
\cmidrule{2-11}
& DMGCRN & \textbf{2.61} & \textbf{5.01}  & \textbf{6.72} & \textbf{2.99}  & \textbf{6.04} & \textbf{8.20} & \textbf{3.38} & \textbf{7.07} & \textbf{9.91}\\
\bottomrule
\end{tabular}
}
\end{table*}
\begin{table*}[t]
\aboverulesep=0ex
\belowrulesep=0ex
\centering
\caption{Traffic flow prediction performance comparison of DMGCRN and baselines on PEMS03 and PEMS05. }
\label{pems}
\resizebox{\textwidth}{!}{
\begin{tabular}{c|l|ccc|ccc|ccc} 
\toprule
\multirow{2}{*}{Datasets} & \multirow{2}{*}{Methods} & \multicolumn{3}{c|}{15 min} & \multicolumn{3}{c|}{30 min} & \multicolumn{3}{c}{1 hour} \\
\cmidrule{3-11}
    &  & MAE & RMSE & MAPE (\%) & MAE & RMSE & MAPE (\%) & MAE & RMSE & MAPE (\%) \\
\midrule
\multirow{10}{*}{PEMS03}
& HA & 22.19 & 34.42 & 20.78 & 26.12 & 40.51 & 24.45 & 36.94 & 56.61 & 35.57\\
& SVR & 17.01 & 28.25 & 18.96 & 20.69 & 33.16 & 22.41 & 28.48 & 43.24 & 31.93\\
& FC-LSTM & 16.01 & 26.13 & 15.51 & 19.19 & 30.97 & 18.54 & 26.04 & 40.74 & 25.88\\
\cmidrule{2-11}
& DCRNN & 14.93 & 26.09 & 15.27 & 16.58 & 28.91 & 16.32 & 19.59 & 33.39 & 19.00\\
& STGCN & 14.75 & 25.93 & 14.84 & 16.72 & 28.29 & 16.31 & 20.34 & 33.91 & 19.01\\
\cmidrule{2-11}
& Graph WaveNet & 14.73 & 25.80 & 14.78 & 15.67 & 27.96 & 16.02 & 17.58 & 29.85 & 18.17\\
& AGCRN & 14.75 & 26.16 & 14.92 & 15.87 & 27.97 & 15.94 & 17.66 & 30.61 & 18.47\\
& STFGNN & 15.18 & 26.04 & 15.02 & 16.59  & 28.37 & 16.00 & 19.21 & 32.34 & 18.15\\
\cmidrule{2-11}
& GMAN & 15.38 & 25.67 & 15.20 & 16.05 & 26.80 & 16.76 & 17.51 & 28.98 & 18.32\\
\cmidrule{2-11}
& DMGCRN & \textbf{14.53} & \textbf{24.79} & \textbf{14.68} & \textbf{15.09} & \textbf{26.39} & \textbf{15.20} & \textbf{16.98} & \textbf{28.07} & \textbf{17.11}\\
\midrule
\multirow{10}{*}{PEMS05}
& HA & 10.60 & 16.48  & 33.92 & 12.13 & 18.82 & 38.55 & 16.66 & 25.51 & 53.54\\
& SVR & 8.75 & 13.36 & 28.90 & 10.09 & 15.30 & 32.34 & 13.37 & 15.96 & 45.33\\
& FC-LSTM & 8.67 & 13.74 & 27.83 & 9.76 & 15.50 & 31.12 & 12.41 & 19.54 & 39.46\\
\cmidrule{2-11}
& DCRNN & 7.59 & 12.25 & 26.65 & 7.96 & 12.93 & 28.72 & 8.75 & 14.29 & 32.54\\
& STGCN & 7.54 & 12.08 & 27.75 & 8.43 & 13.63 & 29.95 & 10.12 & 16.63 & 34.07\\
\cmidrule{2-11}
& Graph WaveNet & 7.16 & 11.73 & 25.46 & 7.65 & 12.74 & 26.77 & 8.15 & 13.98 & 28.91\\
& AGCRN & 7.15 & 11.60 & 24.97 & 7.53 & 12.81 & 26.69 & 8.49 & 13.51 & 28.22\\
& STFGNN & 8.13 & 13.17 & 27.93 & 8.63 & 14.18 & 29.84 & 9.75 & 16.47 & 30.67\\
\cmidrule{2-11}
& GMAN & 7.43 & 12.55  & 26.58 & 7.55 & 12.81 & 28.19 & 8.04 & 13.57 & 32.03\\
\cmidrule{2-11}
& DMGCRN & \textbf{6.72} & \textbf{11.08} & \textbf{24.05} & \textbf{7.23}  & \textbf{12.32} & \textbf{26.05} & \textbf{7.71} & \textbf{13.20} & \textbf{27.55}\\
\bottomrule
\end{tabular}
}
\end{table*}
\begin{figure*}[t]
\centering
    \begin{subfigure}{0.33\linewidth}
    \includegraphics[width=\linewidth]{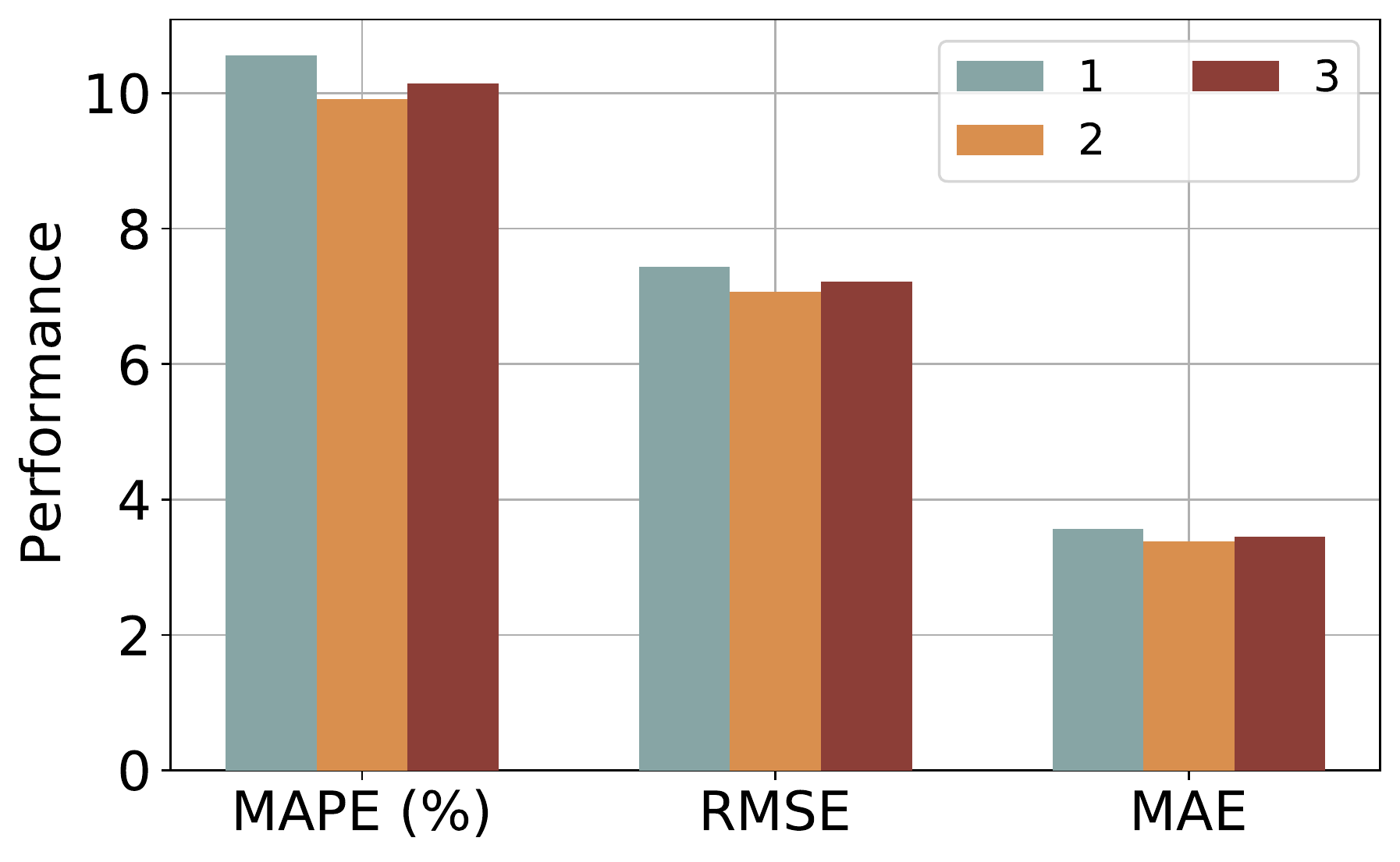}
    \caption{Layers.}
    \label{layer}
  \end{subfigure}%
  \begin{subfigure}{0.33\linewidth}
    \includegraphics[width=\linewidth]{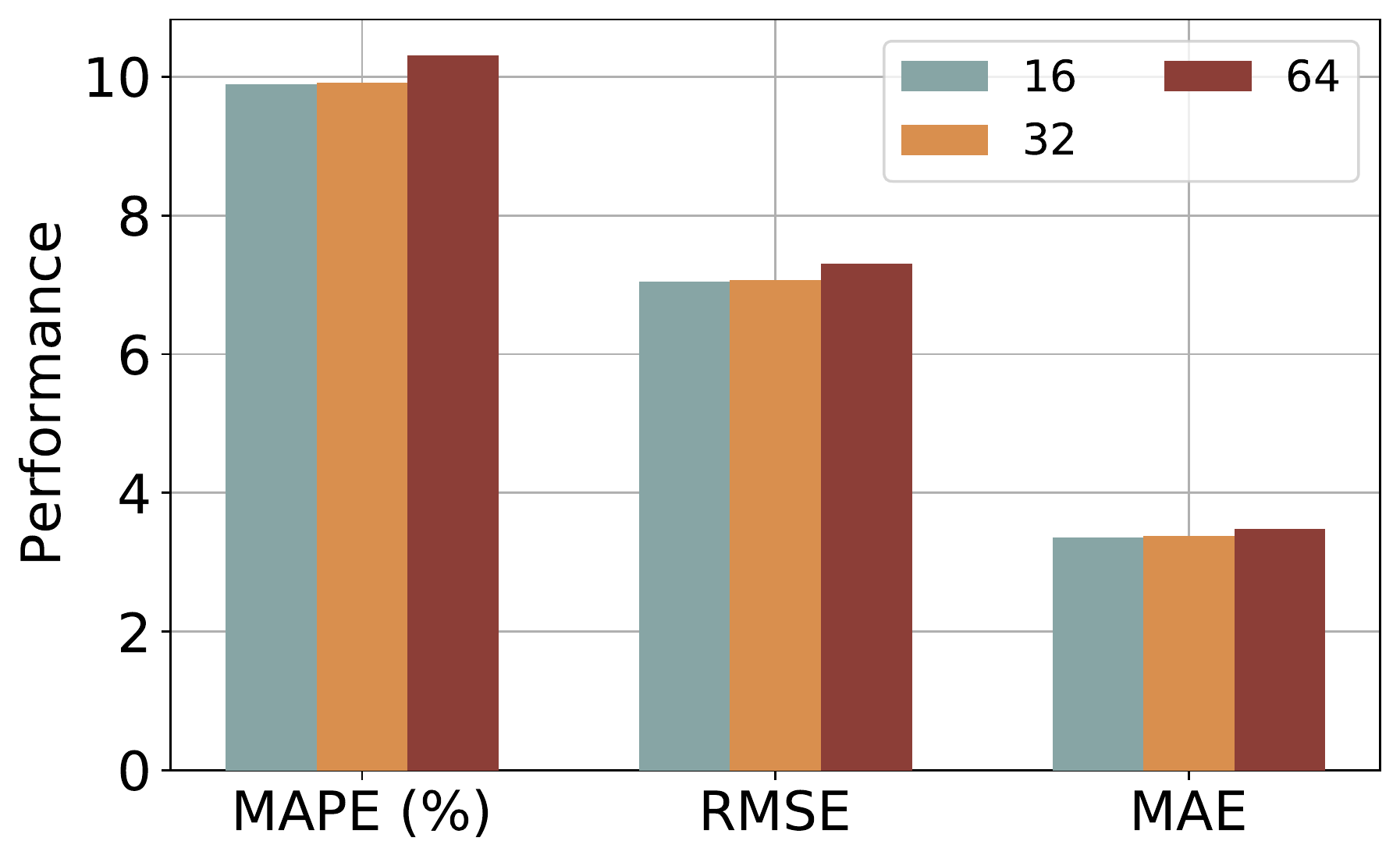}
    \caption{Batchsize.}
    \label{bs}
  \end{subfigure}%
  \begin{subfigure}{0.33\linewidth}
    \includegraphics[width=\linewidth]{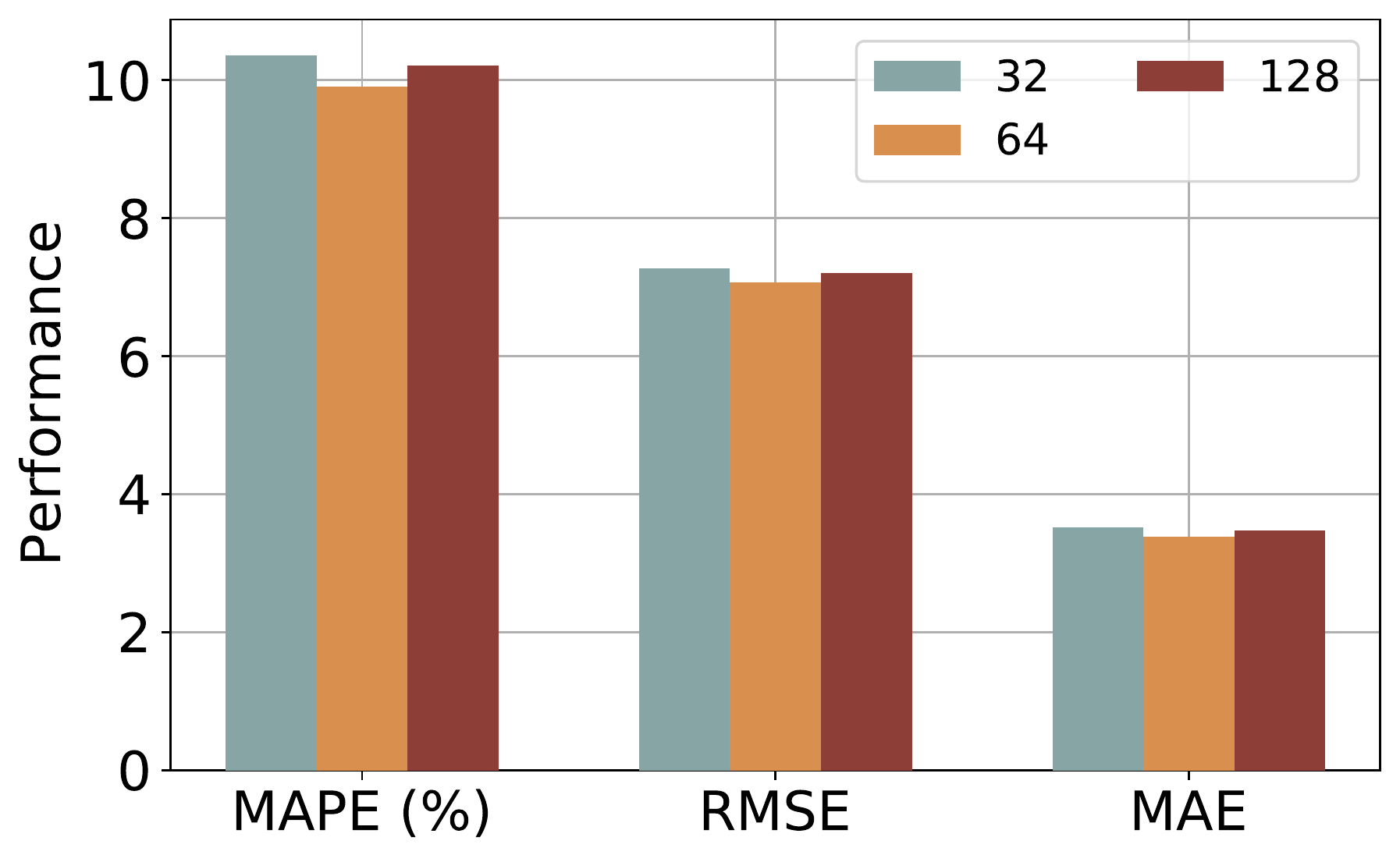}
    \caption{Dimensions.}
    \label{dim}
  \end{subfigure}%
  \caption{Impact of hyperparameter settings.}
  \label{paramshow}
\end{figure*}
\begin{figure*}[t]
\centering
    \begin{subfigure}{0.33\linewidth}
    \includegraphics[width=\linewidth]{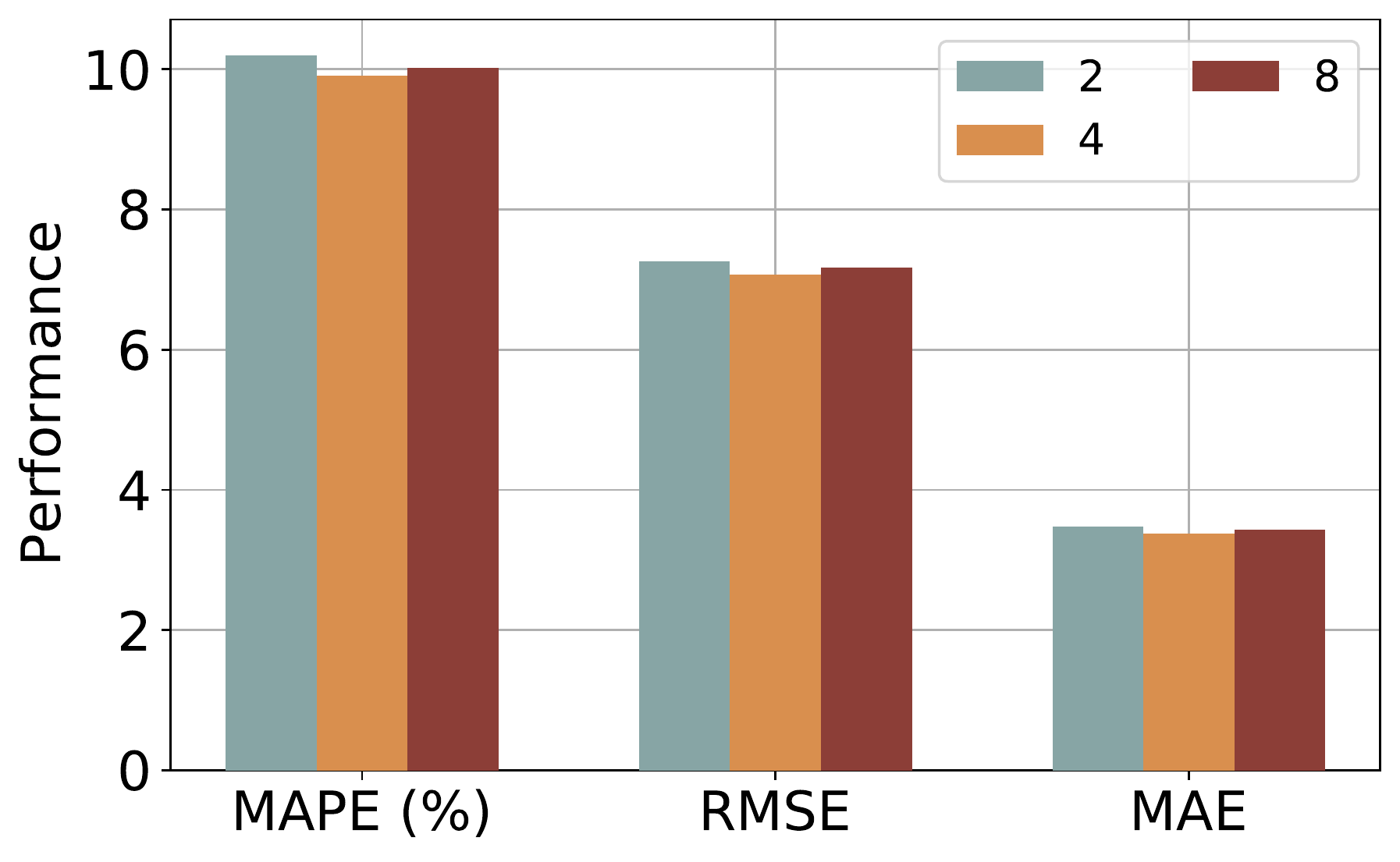}
    \caption{Effectiveness of the number of regions.}
    \label{ablnum}
  \end{subfigure}%
   \begin{subfigure}{0.33\linewidth}
    \includegraphics[width=\linewidth]{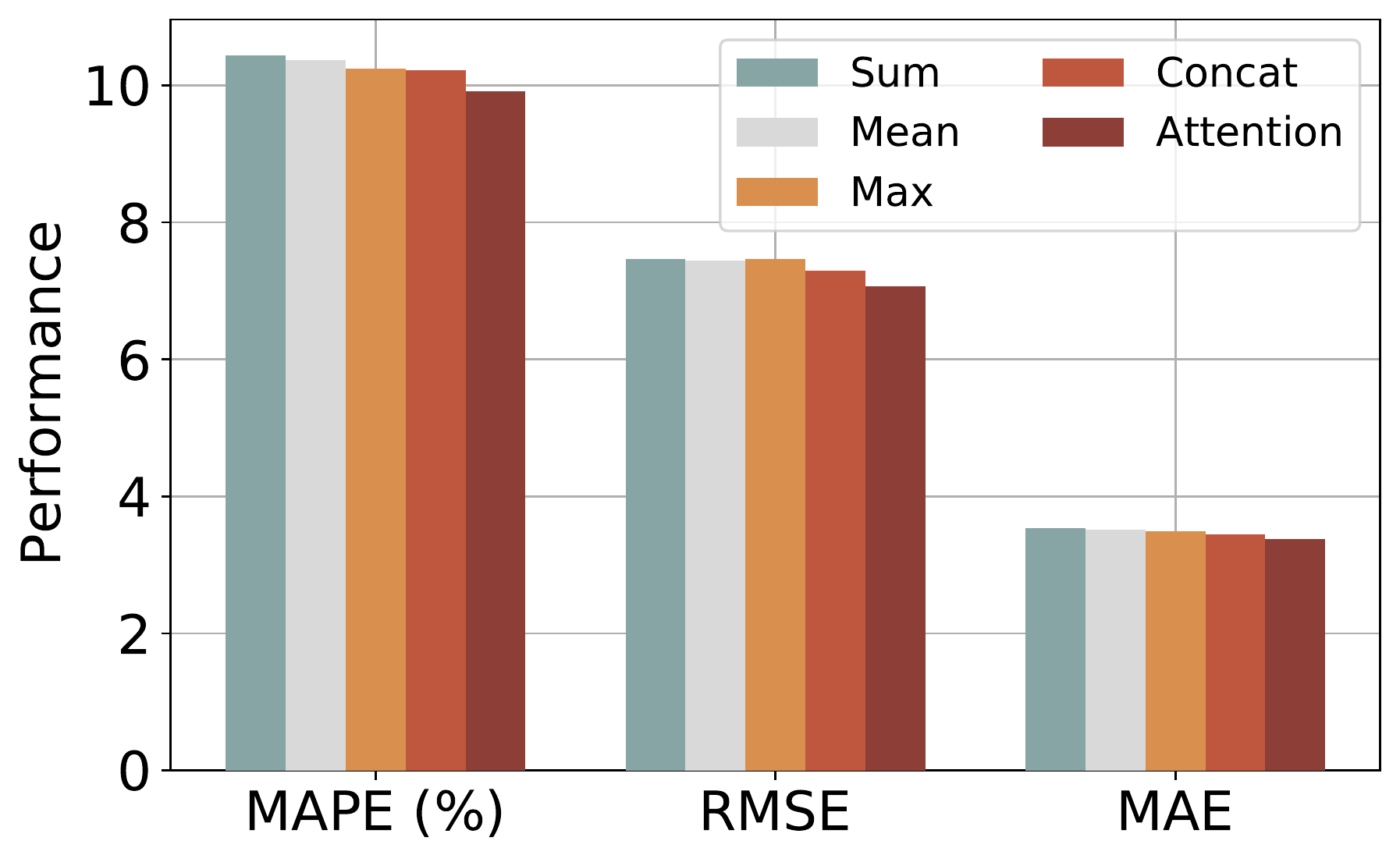}
    \caption{Effectiveness of the attention.}
    \label{ablatt}
  \end{subfigure}%
  \begin{subfigure}{0.33\linewidth}
    \includegraphics[width=\linewidth]{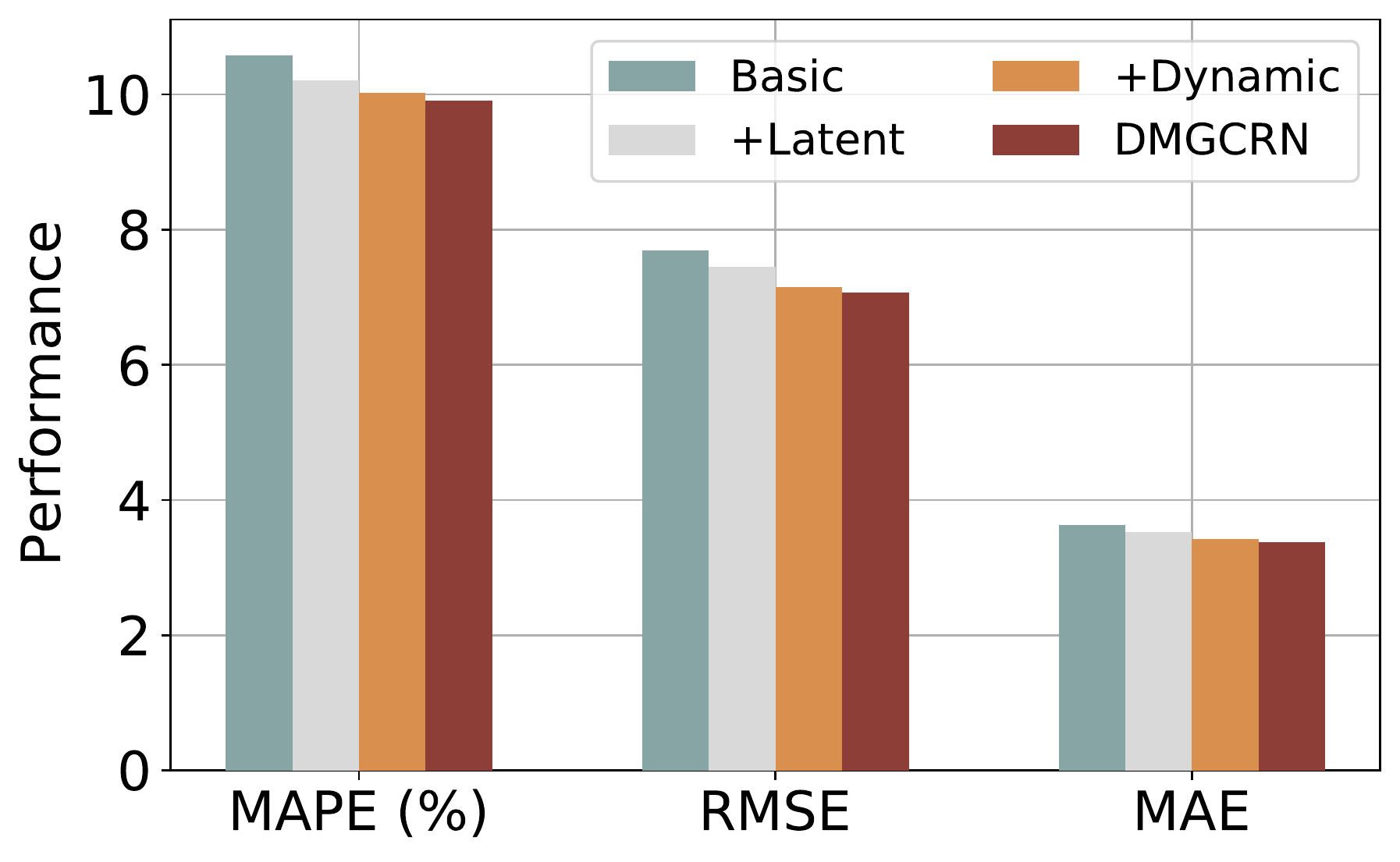}
    \caption{Effectiveness of the dynamic multi-graph.}
    \label{abldmg}
  \end{subfigure}%
  \caption{Ablation study.}
  \label{ablshow}
\end{figure*}
\subsection{Experiment Results}
Table \ref{metr} and Table \ref{pems} show the comparison of different models on traffic speed and flow forecasting. METR-LA and PEMS03 are widely used for traffic forecasting performance evaluation. We observe DMGCRN achieves state-of-the-art results on all the tasks. In the following, we discuss experimental results of traffic forecasting.

SVR merely captures periodic patterns of traffic data but ignores recent temporal correlations, while HA is just the opposite. We can easily observe that several deep learning-based models (FC-LSTM, DCRNN, Graph WaveNet, and so on) usually out-performance traditional statistical methods (HA) and shallow machine learning methods (SVR). Recently, other researchers provide some deep learning models, such as LSTM, to capture non-linear temporal dependence. They achieve insignificant improvement compared with the traditional methods because of the limitation of employing temporal correlations only. STGCN combines GCN and 1D-CNN to capture spatio-temporal correlations in traffic data, respectively. We think that performing STGCN on the METR-LA dataset is not satisfactory because 1D-CNN is difficult to capture the long-term temporal correlations. DCRNN uses DCN and RNN to capture spatial dependence and long short-term temporal dependencies and achieves a significant improvement in the three datasets compared with the previous methods. However, DCRNN and STGCN lack of consideration for the complex and dynamic conditions of the actual road network because they adopt the fixed road network structure as the prior graph adjacency matrix. Graph WaveNet and AGCRN propose to learn an adaptive adjacency matrix through back-propagation and make improvements to STGCN and DCRNN. Although the adaptive adjacency matrix is more complete and accurate than the road network based adjacency matrix, but they are limited by over-fitting, just like performing AGCRN on the METR-LA dataset. STFGNN proposes a novel temporal graph mechanism because they believe that sensors with the similar temporal pattern are highly close. We consider that when the multivariate time series data are relatively similar, the temporal graph will not be effective, just as they cannot achieve better results in traffic speed prediction. Besides, they use 1D-CNN to extract temporal correlations, thus ignore long-term correlations result in worse performance of one hour forecasting. GMAN uses the self-attention mechanism to capture spatio-temporal dependencies and make significant improvements in long-term prediction tasks. We noticed that the self-attention mechanism is not sensitive to local information. Therefore, GMAN is not effective in short-term prediction tasks. Moreover, the self-attention mechanism is limited by the quadratic problem, which requires abundant computing resources and cannot train efficiently.

DMGCRN proposes not only a novel latent graph to mine spatio-temporal correlations of sensors with similar structure but also a novel dynamic GCN mechanism to coarse-grained dynamic modeling spatial region correlations. Therefore, DMGCRN has achieved better results on more complex datasets of the road network, \emph{i.e.}, the results of DMGCRN on METR-LA and PEMS03 are better than PEMS05. Besides, DMGCRN integrates the DMGCN into GRU to capture long short-term temporal dependencies, thus DMGCRN has the best performance in long short-term traffic forecasts on all tasks.
\subsection{Evaluation on Hyperparameter Settings}
We tune hyperparameters on the validation set of all datasets and show the results of DMGCRN on the test set of METR-LA dataset under different parameters in Figure \ref{paramshow}. As shown in Figure \ref{layer} and \ref{dim}, we observe that increasing layers and dimensions of DMGCRN initially lowers the performance but then encounters over-fitting problem when the model becomes too complex. From Figure \ref{bs}, we find that our model performance is better as the batch size decreases.
\subsection{Ablation Study}
\subsubsection{Evaluation on the Number of Regions}
The distance graph can naturally use the latitude and longitude to divide the neighbors of each sensor into four different regions, but the number of regions of the latent graph determined according to the dimension of latent representation of sensors, \emph{i.e.}, if the dimension of latent representation of sensors is $r$, there are $2^r$ regions. We conduct experiments of region division. According to the experimental results on METR-LA dataset shown in Figure \ref{ablnum}, the effect of dividing four regions is the best and the most practical choice.
\subsubsection{Evaluation on Attention}
In order to investigate the effect of the attention operation in our model, we conduct experiments on METR-LA dataset. As shown is Figure \ref{ablatt}, the attention method is outperforms other methods. We consider that the attention operation can consider all regions, unlike max method to lose some regions information. The spatial dependence change dynamically over time, so we use attention method to calculate the weight of each region at different time steps instead of directly summing or averaging.
\begin{figure*}[t]
\centering
    \begin{subfigure}{0.24\linewidth}
    \includegraphics[width=\linewidth,height=2.5cm]{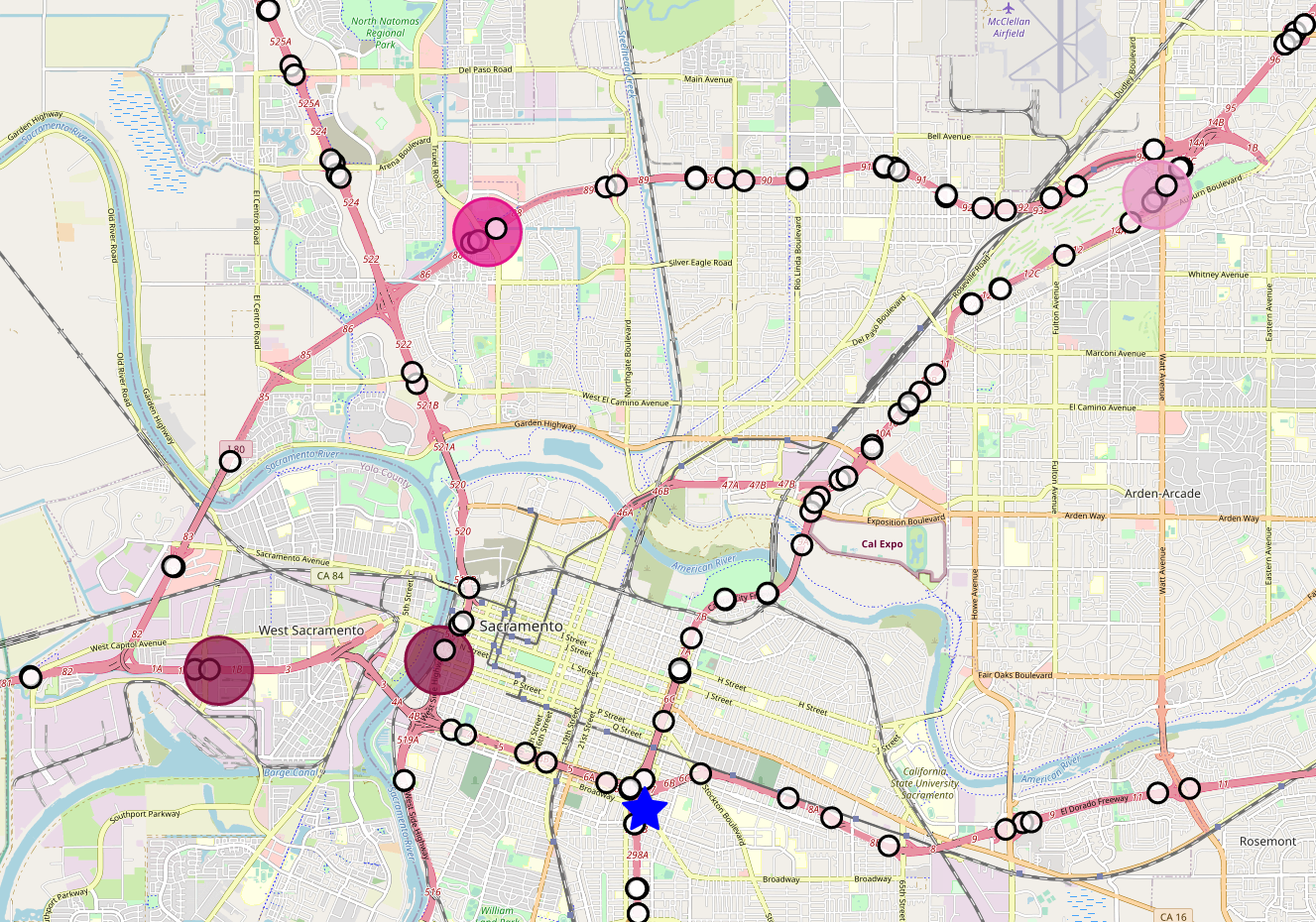}
    \caption{Latent graph on weekday.}
  \end{subfigure}\hfill
   \begin{subfigure}{0.24\linewidth}
    \includegraphics[width=\linewidth,height=2.5cm]{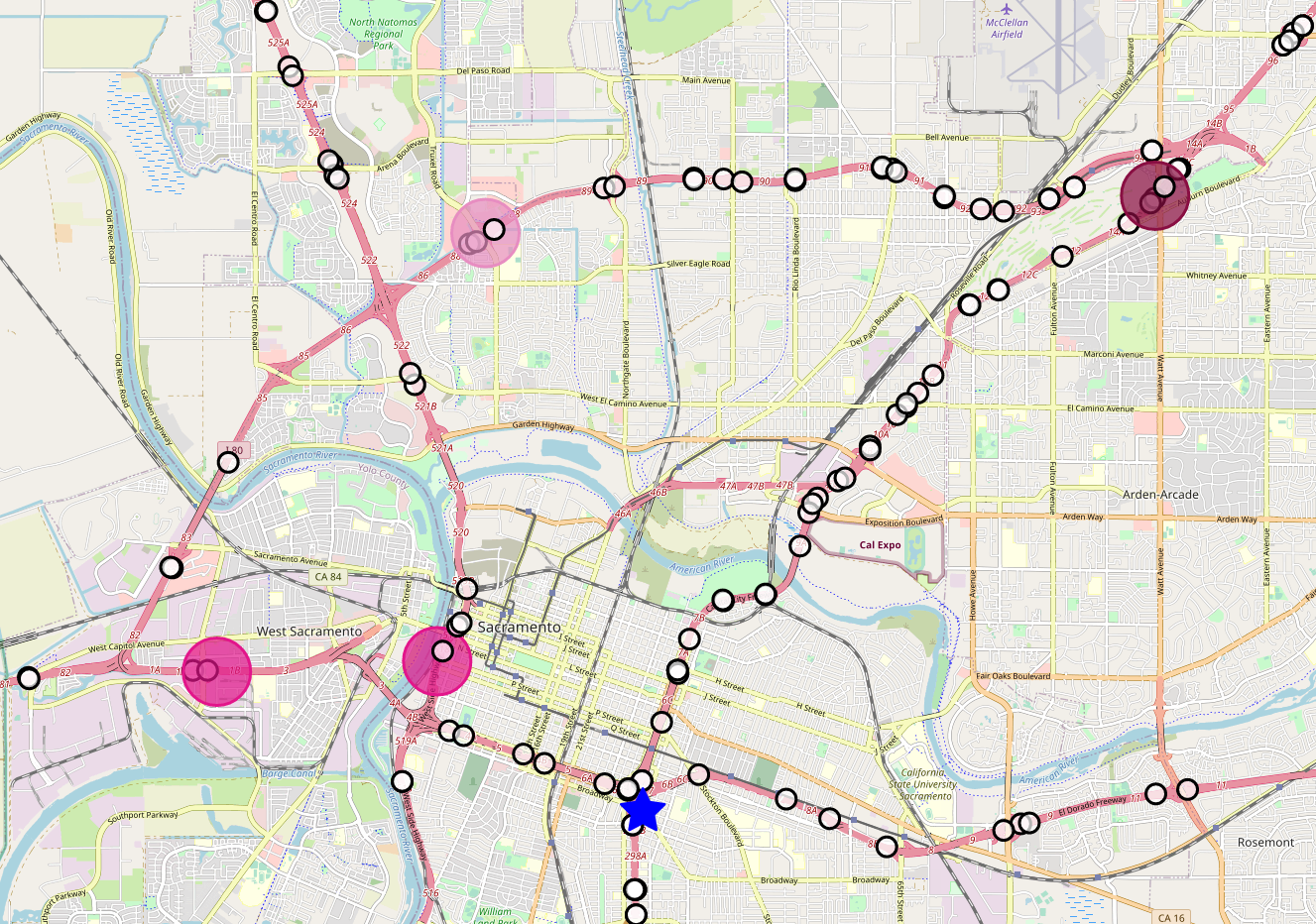}
    \caption{Latent graph on weekends.}
  \end{subfigure}\hfill
  \begin{subfigure}{0.24\linewidth}
    \includegraphics[width=\linewidth,height=2.5cm]{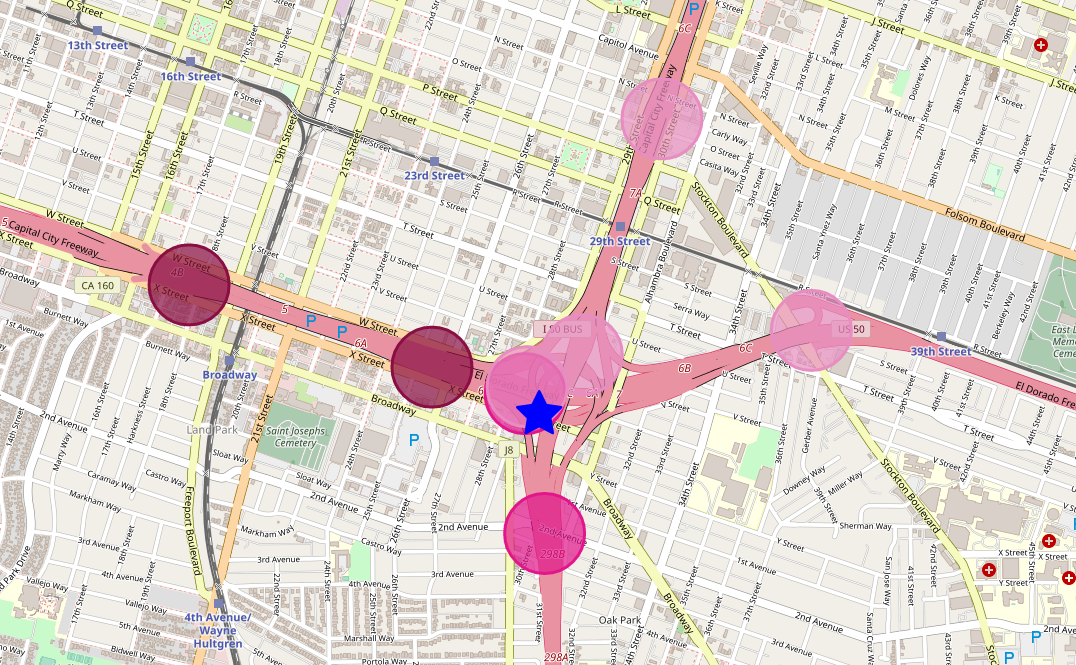}
    \caption{Distance graph on weekday.}
  \end{subfigure}\hfill
   \begin{subfigure}{0.24\linewidth}
    \includegraphics[width=\linewidth,height=2.5cm]{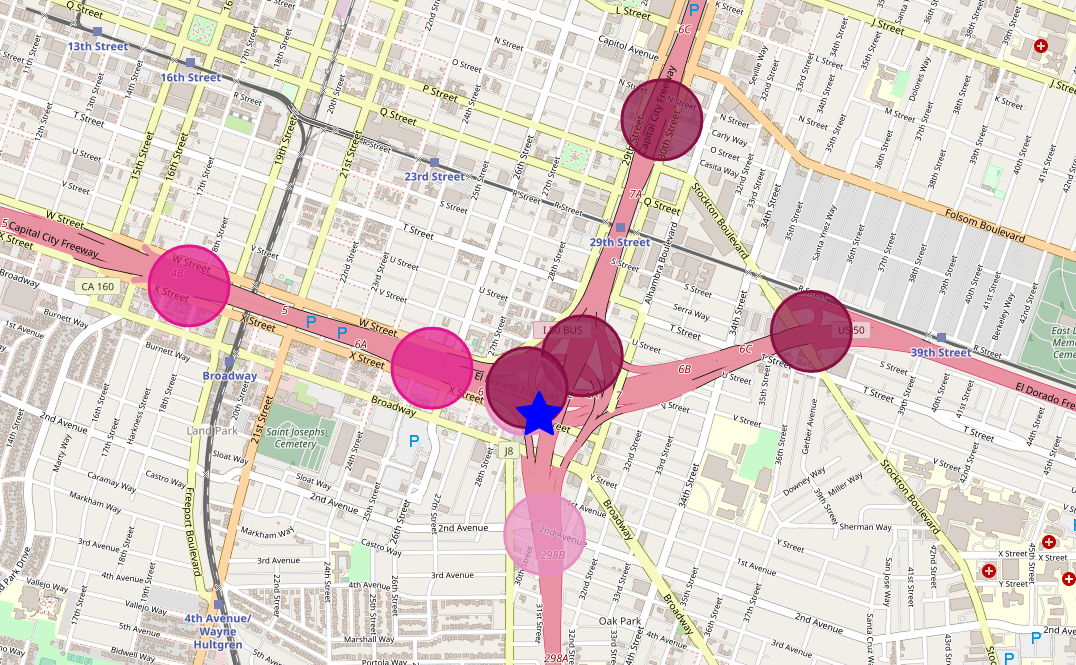}
    \caption{Distance graph on weekends.}
  \end{subfigure}
  \caption{Maps neighbors of the sensor 182 in the latent graph and the distance graph on the PEMS03 datatset. Blue star is the sensor 182. Red circles are neighbors of sensor 182, and the darker the color, the greater the attention weight. Besides, the same shade of red circles are in the same region.}
  \label{vis}
\end{figure*}
\subsubsection{Ablation Experiments}
To verify the different parts in DMGCRN, we propose three variants and compare them with DMGCRN for ablation experiments on METR-LA dataset. These variants follow the setup of DMGCRN. The predicted value of the next hour is shown in Figure \ref{abldmg}. The detailed description of three variants is described as below:
\begin{itemize}
    \item Basic: This model replaces the linear unit in the GRU with graph convolution, and the adjacency matrix does not equip with mask matrix. Meanwhile, we use an encoder-decoder architecture for traffic forecasting.
    \item $+$Latent: This method uses dual-channel graph convolution to capture neighboring spatial correlations and structural spatial correlations through the distance graph and the latent graph. Besides, we also utilize the fusion module in this method as in DMGCRN.
    \item $+$Dynamic: Based on the $+$Latent model, we divide neighbors of each sensor into four regions and perform dynamic graph convolution according to the method in Section \ref{DGCN}, except using the mask matrix.
    \item DMGCRN: This model adds mask matrix based on the $+$Dynamic model.
\end{itemize}
As Figure \ref{abldmg} illustrates, $+$Latent is significantly better than the Basic model. Because the graph convolution based on the latent graph adjacency matrix can capture the remote spatial dependence efficiently. Different regions at different times should have different effects for sensors in the road network. However, the $+$Latent model ignores to dynamically capture spatial dependence. $+$Dynamic aggregates information from different regions dynamically and outperforms the $+$Latent by a large margin in MAE, RMSE, and MAPE metric. Besides, the mask matrix is a learnable matrix, which tunes the weights adaptively between each sensor and its neighbors through back-propagation and improves the forecasting performance.
\subsection{Visualization}
Figure \ref{vis} visualizes neighbors of sensor 182 in the latent graph and the distance graph on the PEMS03 dataset. The location of neighbors of sensor 128 in the latent graph are like itself at intersections, and their locations in the city have similar functions. Therefore, we use sensors with similar spatio-temporal patterns through the latent graph to assist in more accurate traffic predictions. Besides, we find the weights of neighbors in different regions are different during the weekday and the weekend, which proves the correctness of dynamically modeling of the hierarchical spatial dependence.
\section{Conclusion}
In this paper, we propose a novel dynamic multi-graph convolution recurrent network for traffic forecasting. The dynamic multi-graph convolution recurrent network integrates the DMGCN into GRU to capture spatio-temporal dependencies simultaneously. In the DMGCN, we construct the latent graph and the distance graph to capture the neighboring and remote spatial correlations individually. Then neighbors of each sensor in the latent graph and the distance graph are divided into four regions corresponding to the relative relationships, respectively. Finally, we use GCN and attention to dynamically capture the hierarchical spatial dependence of sensors and regions. The experiment results on the METR-LA, PEMS03, and PEMS05 datasets show DMGCRN achieves the state-of-the-art performance. In the future, we will extension the proposed model for more general spatio-temporal forecasting tasks such as the stock price prediction.

%



\section*{Acknowledgment}
Many thanks to the Beijing Key Laboratory of Mobile Computing and Pervasive Device, Institute of Computing Technology Chinese Academy of Sciences.




\bibliographystyle{IEEEtran}
\bibliography{HEFGAT}

\begin{thebibliography}{10}
\providecommand{\url}[1]{#1}
\csname url@samestyle\endcsname
\providecommand{\newblock}{\relax}
\providecommand{\bibinfo}[2]{#2}
\providecommand{\BIBentrySTDinterwordspacing}{\spaceskip=0pt\relax}
\providecommand{\BIBentryALTinterwordstretchfactor}{4}
\providecommand{\BIBentryALTinterwordspacing}{\spaceskip=\fontdimen2\font plus
\BIBentryALTinterwordstretchfactor\fontdimen3\font minus
  \fontdimen4\font\relax}
\providecommand{\BIBforeignlanguage}[2]{{%
\expandafter\ifx\csname l@#1\endcsname\relax
\typeout{** WARNING: IEEEtran.bst: No hyphenation pattern has been}%
\typeout{** loaded for the language `#1'. Using the pattern for}%
\typeout{** the default language instead.}%
\else
\language=\csname l@#1\endcsname
\fi
#2}}
\providecommand{\BIBdecl}{\relax}
\BIBdecl

\bibitem{dimitrakopoulos2010intelligent}
G.~Dimitrakopoulos and P.~Demestichas, ``Intelligent transportation systems,''
  \emph{IEEE Vehicular Technology Magazine}, vol.~5, no.~1, pp. 77--84, 2010.

\bibitem{wu2016short}
Y.~Wu and H.~Tan, ``Short-term traffic flow forecasting with spatial-temporal
  correlation in a hybrid deep learning framework,'' \emph{arXiv preprint
  arXiv:1612.01022}, 2016.

\bibitem{DBLP:conf/nips/CaoWDZZHTXBTZ20}
D.~Cao, Y.~Wang, J.~Duan, C.~Zhang, X.~Zhu, C.~Huang, Y.~Tong, B.~Xu, J.~Bai,
  J.~Tong, and Q.~Zhang, ``Spectral temporal graph neural network for
  multivariate time-series forecasting,'' in \emph{Advances in Neural
  Information Processing Systems 33: Annual Conference on Neural Information
  Processing Systems, NeurIPS, December 6-12, virtual}, 2020.

\bibitem{williams2003modeling}
B.~M. Williams and L.~A. Hoel, ``Modeling and forecasting vehicular traffic
  flow as a seasonal arima process: Theoretical basis and empirical results,''
  \emph{Journal of transportation engineering}, vol. 129, no.~6, pp. 664--672,
  2003.

\bibitem{lu2016integrating}
Z.~Lu, C.~Zhou, J.~Wu, H.~Jiang, and S.~Cui, ``Integrating granger causality
  and vector auto-regression for traffic prediction of large-scale wlans,''
  \emph{KSII Transactions on Internet and Information Systems (TIIS)}, vol.~10,
  no.~1, pp. 136--151, 2016.

\bibitem{wu2004travel}
C.-H. Wu, J.-M. Ho, and D.-T. Lee, ``Travel-time prediction with support vector
  regression,'' \emph{IEEE transactions on intelligent transportation systems},
  vol.~5, no.~4, pp. 276--281, 2004.

\bibitem{van2012short}
J.~Van~Lint and C.~Van~Hinsbergen, ``Short-term traffic and travel time
  prediction models,'' \emph{Artificial Intelligence Applications to Critical
  Transportation Issues}, vol.~22, no.~1, pp. 22--41, 2012.

\bibitem{chung2014empirical}
J.~Chung, C.~Gulcehre, K.~Cho, and Y.~Bengio, ``Empirical evaluation of gated
  recurrent neural networks on sequence modeling,'' \emph{arXiv preprint
  arXiv:1412.3555}, 2014.

\bibitem{lv2018lc}
Z.~Lv, J.~Xu, K.~Zheng, H.~Yin, P.~Zhao, and X.~Zhou, ``Lc-rnn: A deep learning
  model for traffic speed prediction.'' in \emph{IJCAI}, 2018, pp. 3470--3476.

\bibitem{DBLP:conf/ssst/ChoMBB14}
K.~Cho, B.~van Merrienboer, D.~Bahdanau, and Y.~Bengio, ``On the properties of
  neural machine translation: Encoder-decoder approaches,'' in
  \emph{Proceedings of SSST@EMNLP 2014, Eighth Workshop on Syntax, Semantics
  and Structure in Statistical Translation, Doha, Qatar, 25 October}, 2014, pp.
  103--111.

\bibitem{yao2018deep}
H.~Yao, F.~Wu, J.~Ke, X.~Tang, Y.~Jia, S.~Lu, P.~Gong, J.~Ye, and Z.~Li, ``Deep
  multi-view spatial-temporal network for taxi demand prediction,'' in
  \emph{Proceedings of the AAAI Conference on Artificial Intelligence},
  vol.~32, no.~1, 2018.

\bibitem{cui2018deep}
Z.~Cui, R.~Ke, Z.~Pu, and Y.~Wang, ``Deep bidirectional and unidirectional lstm
  recurrent neural network for network-wide traffic speed prediction,''
  \emph{arXiv preprint arXiv:1801.02143}, 2018.

\bibitem{wan2019multivariate}
R.~Wan, S.~Mei, J.~Wang, M.~Liu, and F.~Yang, ``Multivariate temporal
  convolutional network: A deep neural networks approach for multivariate time
  series forecasting,'' \emph{Electronics}, vol.~8, no.~8, p. 876, 2019.

\bibitem{zhang2016dnn}
J.~Zhang, Y.~Zheng, D.~Qi, R.~Li, and X.~Yi, ``Dnn-based prediction model for
  spatio-temporal data,'' in \emph{Proceedings of the 24th ACM SIGSPATIAL
  International Conference on Advances in Geographic Information Systems},
  2016, pp. 1--4.

\bibitem{zhang2017deep}
J.~Zhang, Y.~Zheng, and D.~Qi, ``Deep spatio-temporal residual networks for
  citywide crowd flows prediction,'' in \emph{Proceedings of the AAAI
  Conference on Artificial Intelligence}, vol.~31, no.~1, 2017.

\bibitem{DBLP:conf/iclr/KipfW17}
T.~N. Kipf and M.~Welling, ``Semi-supervised classification with graph
  convolutional networks,'' in \emph{5th International Conference on Learning
  Representations, {ICLR} 2017, Toulon, France, April 24-26, 2017, Conference
  Track Proceedings}, 2017.

\bibitem{DBLP:conf/nips/DefferrardBV16}
M.~Defferrard, X.~Bresson, and P.~Vandergheynst, ``Convolutional neural
  networks on graphs with fast localized spectral filtering,'' in
  \emph{Advances in Neural Information Processing Systems 29: Annual Conference
  on Neural Information Processing Systems 2016, December 5-10, 2016,
  Barcelona, Spain}, 2016, pp. 3837--3845.

\bibitem{DBLP:conf/iclr/LiYS018}
Y.~Li, R.~Yu, C.~Shahabi, and Y.~Liu, ``Diffusion convolutional recurrent
  neural network: Data-driven traffic forecasting,'' in \emph{6th International
  Conference on Learning Representations, {ICLR}, Vancouver, BC, Canada, April
  30 - May 3, Conference Track Proceedings}, 2018.

\bibitem{DBLP:conf/ijcai/YuYZ18}
B.~Yu, H.~Yin, and Z.~Zhu, ``Spatio-temporal graph convolutional networks: {A}
  deep learning framework for traffic forecasting,'' in \emph{Proceedings of
  the Twenty-Seventh International Joint Conference on Artificial Intelligence,
  {IJCAI} , July 13-19, Stockholm, Sweden}, 2018, pp. 3634--3640.

\bibitem{lv2020temporal}
M.~Lv, Z.~Hong, L.~Chen, T.~Chen, T.~Zhu, and S.~Ji, ``Temporal multi-graph
  convolutional network for traffic flow prediction,'' \emph{IEEE Transactions
  on Intelligent Transportation Systems}, 2020.

\bibitem{DBLP:conf/aaai/LiZ21}
M.~Li and Z.~Zhu, ``Spatial-temporal fusion graph neural networks for traffic
  flow forecasting,'' in \emph{Thirty-Fifth {AAAI} Conference on Artificial
  Intelligence, {AAAI} , Virtual Event, February 2-9}, 2021, pp. 4189--4196.

\bibitem{fang2021spatial}
Z.~Fang, Q.~Long, G.~Song, and K.~Xie, ``Spatial-temporal graph ode networks
  for traffic flow forecasting,'' in \emph{Proceedings of the 27th ACM SIGKDD
  Conference on Knowledge Discovery \& Data Mining}, 2021, pp. 364--373.

\bibitem{berndt1994using}
D.~J. Berndt and J.~Clifford, ``Using dynamic time warping to find patterns in
  time series.'' in \emph{KDD workshop}, vol.~10, no.~16.\hskip 1em plus 0.5em
  minus 0.4em\relax Seattle, WA, USA:, 1994, pp. 359--370.

\bibitem{zheng2020gman}
C.~Zheng, X.~Fan, C.~Wang, and J.~Qi, ``Gman: A graph multi-attention network
  for traffic prediction,'' in \emph{Proceedings of the AAAI Conference on
  Artificial Intelligence}, vol.~34, no.~01, 2020, pp. 1234--1241.

\bibitem{chen2020multi}
W.~Chen, L.~Chen, Y.~Xie, W.~Cao, Y.~Gao, and X.~Feng, ``Multi-range attentive
  bicomponent graph convolutional network for traffic forecasting,'' in
  \emph{Proceedings of the AAAI Conference on Artificial Intelligence},
  vol.~34, no.~04, 2020, pp. 3529--3536.

\bibitem{song2020spatial}
C.~Song, Y.~Lin, S.~Guo, and H.~Wan, ``Spatial-temporal synchronous graph
  convolutional networks: A new framework for spatial-temporal network data
  forecasting,'' in \emph{Proceedings of the AAAI Conference on Artificial
  Intelligence}, vol.~34, no.~01, 2020, pp. 914--921.

\bibitem{park2020st}
C.~Park, C.~Lee, H.~Bahng, Y.~Tae, S.~Jin, K.~Kim, S.~Ko, and J.~Choo,
  ``St-grat: A novel spatio-temporal graph attention networks for accurately
  forecasting dynamically changing road speed,'' in \emph{Proceedings of the
  29th ACM International Conference on Information \& Knowledge Management},
  2020, pp. 1215--1224.

\bibitem{guo2021hierarchical}
K.~Guo, Y.~Hu, Y.~Sun, S.~Qian, J.~Gao, and B.~Yin, ``Hierarchical graph
  convolution networks for traffic forecasting,'' in \emph{Proceedings of the
  AAAI Conference on Artificial Intelligence}, vol.~35, no.~1, 2021, pp.
  151--159.

\bibitem{lippi2013short}
M.~Lippi, M.~Bertini, and P.~Frasconi, ``Short-term traffic flow forecasting:
  An experimental comparison of time-series analysis and supervised learning,''
  \emph{IEEE Transactions on Intelligent Transportation Systems}, vol.~14,
  no.~2, pp. 871--882, 2013.

\bibitem{hochreiter1997long}
S.~Hochreiter and J.~Schmidhuber, ``Long short-term memory,'' \emph{Neural
  computation}, vol.~9, no.~8, pp. 1735--1780, 1997.

\bibitem{DBLP:conf/ijcai/WuPLJZ19}
Z.~Wu, S.~Pan, G.~Long, J.~Jiang, and C.~Zhang, ``Graph wavenet for deep
  spatial-temporal graph modeling,'' in \emph{Proceedings of the Twenty-Eighth
  International Joint Conference on Artificial Intelligence, {IJCAI} , Macao,
  China, August 10-16}, 2019, pp. 1907--1913.

\bibitem{DBLP:conf/nips/0001YL0020}
L.~Bai, L.~Yao, C.~Li, X.~Wang, and C.~Wang, ``Adaptive graph convolutional
  recurrent network for traffic forecasting,'' in \emph{Advances in Neural
  Information Processing Systems 33: Annual Conference on Neural Information
  Processing Systems, NeurIPS, December 6-12, virtual}, 2020.

\bibitem{wu2020connecting}
Z.~Wu, S.~Pan, G.~Long, J.~Jiang, X.~Chang, and C.~Zhang, ``Connecting the
  dots: Multivariate time series forecasting with graph neural networks,'' in
  \emph{Proceedings of the 26th ACM SIGKDD International Conference on
  Knowledge Discovery \& Data Mining}, 2020, pp. 753--763.

\bibitem{DBLP:journals/corr/BrunaZSL13}
J.~Bruna, W.~Zaremba, A.~Szlam, and Y.~LeCun, ``Spectral networks and locally
  connected networks on graphs,'' in \emph{2nd International Conference on
  Learning Representations, {ICLR} 2014, Banff, AB, Canada, April 14-16, 2014,
  Conference Track Proceedings}, 2014.

\bibitem{hammond2011wavelets}
D.~K. Hammond, P.~Vandergheynst, and R.~Gribonval, ``Wavelets on graphs via
  spectral graph theory,'' \emph{Applied and Computational Harmonic Analysis},
  vol.~30, no.~2, pp. 129--150, 2011.

\bibitem{atwood2016diffusion}
J.~Atwood and D.~Towsley, ``Diffusion-convolutional neural networks,'' in
  \emph{Advances in neural information processing systems}, 2016, pp.
  1993--2001.

\bibitem{DBLP:conf/nips/HamiltonYL17}
W.~L. Hamilton, Z.~Ying, and J.~Leskovec, ``Inductive representation learning
  on large graphs,'' in \emph{Advances in Neural Information Processing Systems
  30: Annual Conference on Neural Information Processing Systems 2017, December
  4-9, 2017, Long Beach, CA, {USA}}, 2017, pp. 1024--1034.

\bibitem{DBLP:conf/iclr/VelickovicCCRLB18}
P.~Velickovic, G.~Cucurull, A.~Casanova, A.~Romero, P.~Li{\`{o}}, and
  Y.~Bengio, ``Graph attention networks,'' in \emph{6th International
  Conference on Learning Representations, {ICLR}, Vancouver, BC, Canada, April
  30 - May 3, Conference Track Proceedings}, 2018.

\bibitem{DBLP:journals/corr/BahdanauCB14}
D.~Bahdanau, K.~Cho, and Y.~Bengio, ``Neural machine translation by jointly
  learning to align and translate,'' in \emph{3rd International Conference on
  Learning Representations, {ICLR} 2015, San Diego, CA, USA, May 7-9, 2015,
  Conference Track Proceedings}, 2015.

\bibitem{DBLP:conf/emnlp/LuongPM15}
T.~Luong, H.~Pham, and C.~D. Manning, ``Effective approaches to attention-based
  neural machine translation,'' in \emph{Proceedings of the 2015 Conference on
  Empirical Methods in Natural Language Processing, {EMNLP} 2015, Lisbon,
  Portugal, September 17-21, 2015}, pp. 1412--1421.

\bibitem{DBLP:conf/nips/VaswaniSPUJGKP17}
A.~Vaswani, N.~Shazeer, N.~Parmar, J.~Uszkoreit, L.~Jones, A.~N. Gomez,
  L.~Kaiser, and I.~Polosukhin, ``Attention is all you need,'' in
  \emph{Advances in Neural Information Processing Systems 30: Annual Conference
  on Neural Information Processing Systems 2017, December 4-9, 2017, Long
  Beach, CA, {USA}}, 2017, pp. 5998--6008.

\bibitem{liddy2001natural}
E.~D. Liddy, ``Natural language processing,'' 2001.

\bibitem{forsyth2011computer}
D.~Forsyth and J.~Ponce, \emph{Computer vision: A modern approach.}\hskip 1em
  plus 0.5em minus 0.4em\relax Prentice hall, 2011.

\bibitem{dijkstra1959note}
E.~W. Dijkstra \emph{et~al.}, ``A note on two problems in connexion with
  graphs,'' \emph{Numerische mathematik}, vol.~1, no.~1, pp. 269--271, 1959.

\bibitem{chen2020simple}
M.~Chen, Z.~Wei, Z.~Huang, B.~Ding, and Y.~Li, ``Simple and deep graph
  convolutional networks,'' in \emph{International Conference on Machine
  Learning}.\hskip 1em plus 0.5em minus 0.4em\relax PMLR, 2020, pp. 1725--1735.

\bibitem{ribeiro2017struc2vec}
L.~F. Ribeiro, P.~H. Saverese, and D.~R. Figueiredo, ``struc2vec: Learning node
  representations from structural identity,'' in \emph{Proceedings of the 23rd
  ACM SIGKDD international conference on knowledge discovery and data mining},
  2017, pp. 385--394.

\bibitem{DBLP:conf/iclr/PeiWCLY20}
H.~Pei, B.~Wei, K.~C. Chang, Y.~Lei, and B.~Yang, ``Geom-gcn: Geometric graph
  convolutional networks,'' in \emph{8th International Conference on Learning
  Representations, {ICLR}, Addis Ababa, Ethiopia, April 26-30}, 2020.

\bibitem{DBLP:conf/nips/BengioVJS15}
S.~Bengio, O.~Vinyals, N.~Jaitly, and N.~Shazeer, ``Scheduled sampling for
  sequence prediction with recurrent neural networks,'' in \emph{Advances in
  Neural Information Processing Systems 28: Annual Conference on Neural
  Information Processing Systems, December 7-12, Montreal, Quebec, Canada},
  2015, pp. 1171--1179.

\bibitem{sutskever2014sequence}
I.~Sutskever, O.~Vinyals, and Q.~V. Le, ``Sequence to sequence learning with
  neural networks,'' in \emph{Advances in neural information processing
  systems}, 2014, pp. 3104--3112.

\end{thebibliography}
%



%

\vspace{-45pt}
\begin{IEEEbiography}[{\includegraphics[width=1in,height=1.25in,clip,keepaspectratio]{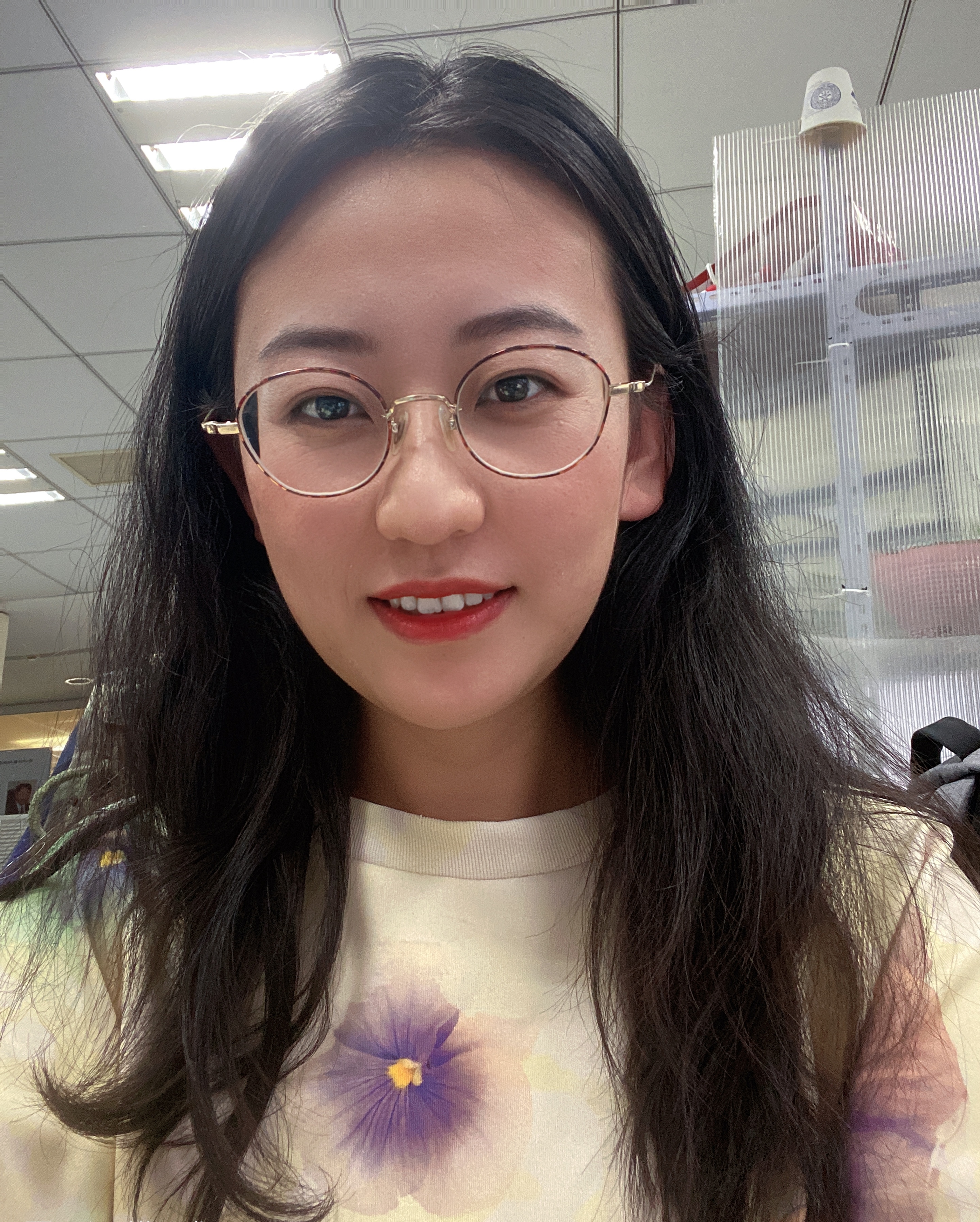}}]{Yanjun Qin}
is currently working toward the Ph.D. degree with the School of Computer Science (National Pilot Software Engineering School), Beijing University of Posts and Telecommunications, Beijing, China. Her current main interests include location based services, pervasive computing, convolution neural networks, and machine learning. He is mainly involved in traffic pattern recognition related project research and implementation.
\end{IEEEbiography}
\vspace{-35pt}
\begin{IEEEbiography}[{\includegraphics[width=1in,height=1.25in,clip,keepaspectratio]{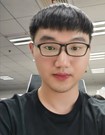}}]{Yuchen Fang}
received the B.S. degree from the School of Information Science and Technology, Beijing Forestry University, Beijing, China. He is currently working toward the M.S. degree with the School of Computer Science (National Pilot Software Engineering School), Beijing University of Posts and Telecommunications, Beijing, China. His current main interests include traffic forecasting and graph neural network.
\end{IEEEbiography}

\vspace{-45pt}
\begin{IEEEbiography}[{\includegraphics[width=1in,height=1.25in,clip,keepaspectratio]{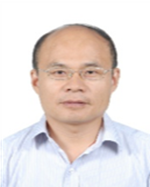}}]{Haiyong Luo}
received the B.S. degree from the Department of Electronics and Information Engineering, Huazhong University of Science and Technology, Wuhan, China, in 1989, the M.S. degree from the School of Information and Communication Engineering, the Beijing University of Posts and Telecommunication China, Beijing, China, in 2002, and Ph.D. degree in computer science from the University of Chines Academy of Sciences, Beijing, China, in 2008. He is currently an Associate Professor with the Institute of Computer Technology, Chinese Academy of Science, Beijing, China. His main research interests are location based services, pervasive computing, mobile computing, and Internet of Things.
\end{IEEEbiography}
\vspace{-45pt}
\begin{IEEEbiography}[{\includegraphics[width=1in,height=1.25in,clip,keepaspectratio]{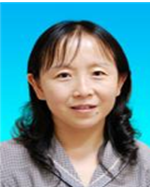}}]{Fang Zhao} received the B.S. degree from the School of Computer Science and Technology, Huazhong University of Science and Technology, Wuan, China, in 1990, and the M.S. and Ph.D. degrees in computer science and technology from the Beijing University of Posts and Telecommunication, Beijing, China, in 2004 and 2009, respectively. She is currently a Professor with the School of Computer Science (National Pilot Software Engineering School), Beijing University of Posts and Telecommunications, Beijing, China. Her research interests include mobile computing, location based services, and computer networks.
\end{IEEEbiography}
\vspace{-45pt}
\begin{IEEEbiography}[{\includegraphics[width=1in,height=1.25in,clip,keepaspectratio]{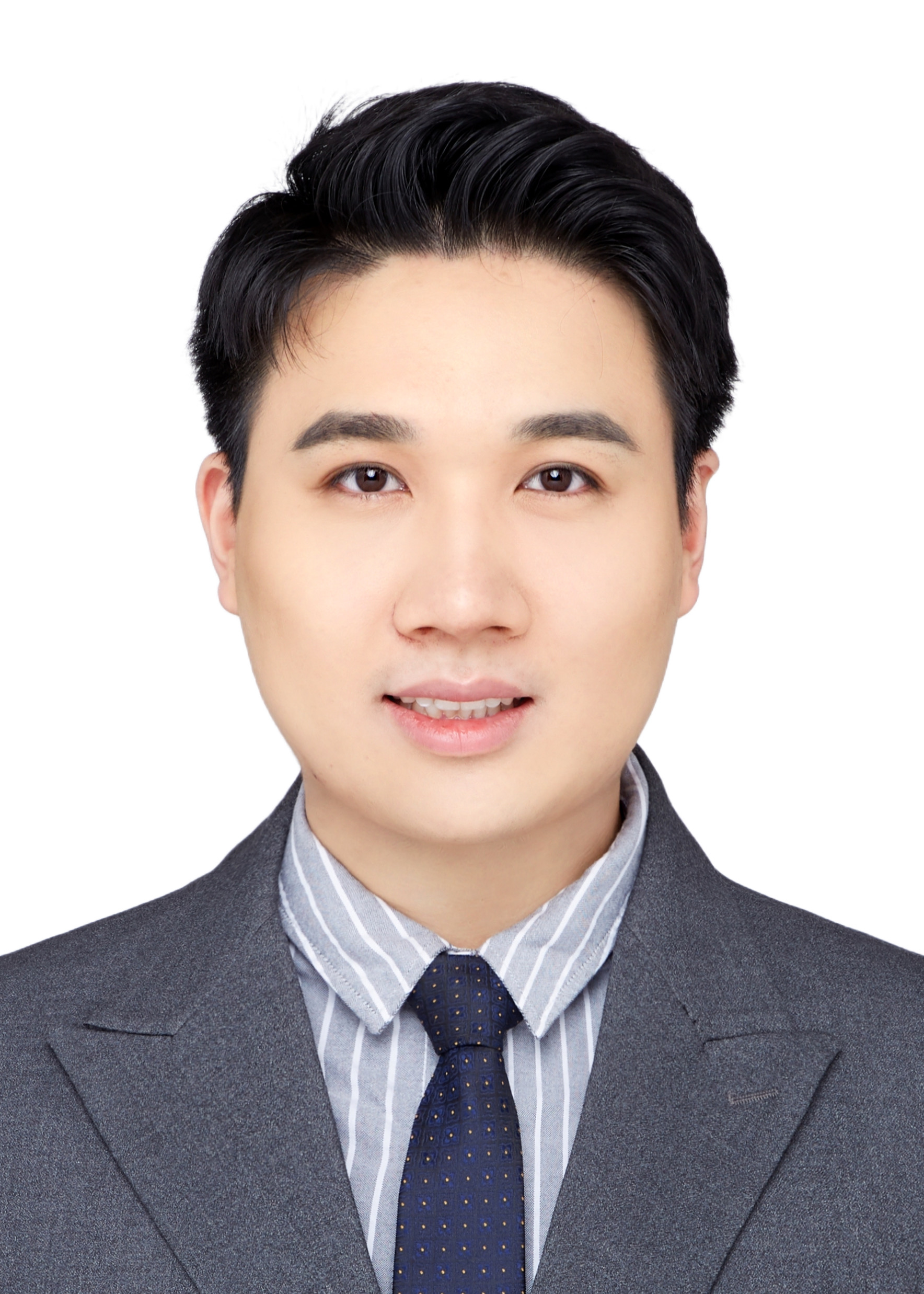}}]{Chenxing Wang}
is currently pursuing the P.h.D with the School of Computer Science (National Pilot Software Engineering School), Beijing University of Posts and Telecommunications, Beijing, China. His current main interests include spatial-temporal data mining, travel time estimation, traffic flow prediction and transportation mode detection using deep learning techniques.
\end{IEEEbiography}
\end{document}